\documentclass[manuscript,screen,nonacm]{acmart}

\AtBeginDocument{%
  }

\usepackage{times}
\usepackage{soul}
\usepackage{url}
\usepackage[utf8]{inputenc}
\usepackage{graphicx}
\usepackage{amsmath}
\usepackage{bbm}
\usepackage{amsthm}
\usepackage{booktabs}
\usepackage[ruled,vlined,linesnumbered]{algorithm2e}
\usepackage{tikz}
\usetikzlibrary{shapes.geometric, arrows, positioning, calc, fit, backgrounds}
\usepackage{multirow}
\usepackage{booktabs}
\usepackage{subcaption}
\usepackage{xcolor}
\usepackage{trimclip}

\tikzstyle{block} = [rectangle, draw, text centered, rounded corners, minimum height=2em, fill=blue!20]
\tikzstyle{input} = [ellipse, draw, text centered, fill=green!20]
\tikzstyle{output} = [ellipse, draw, text centered, fill=red!20]
\tikzstyle{arrow} = [thick,->,>=stealth]

\begin{document}
\title[Preserve my Heritage: Damage Analysis in Cultural Artifacts]{Preserve my Heritage: A Shape-Informed Neural Framework for Damage Analysis and Segmentation in Cultural Artifacts}

\author{\texorpdfstring{\href{https://thiagofreitas.phd}{Thiago Freitas dos Santos}}{Thiago Freitas dos Santos}}
\orcid{0000-0003-3800-7243}
\affiliation{%
  \institution{MIRALab, University of Geneva}
 \city{Geneva}
  \country{Switzerland}
}
\email{thiago.freitasdossantos@unige.ch}

\author{Nadia Magnenat-Thalmann}
\orcid{0000-0002-1459-5960}
\affiliation{%
  \institution{MIRALab, University of Geneva}
  \city{Geneva}
 \country{Switzerland}
}
\email{nadia.thalmann@unige.ch}

\renewcommand{\shortauthors}{Freitas dos Santos et al.}

\begin{abstract}
Automated damage analysis of Cultural Heritage (CH) artifacts is critical for conservation, documentation, and digital reconstruction. However, damage often combines surface degradation with structural loss from missing parts, making its identification from single-view RGB images, the primary visual records in museum and archival collections, challenging. Missing parts alter an artifact's original geometry, while visual similarities among intact surfaces, degraded material, and exposed inner regions complicate damage identification. These challenges are intensified by the scarcity of pixel-level damage annotations. Thus, we propose \emph{DASCH-Net}, a shape-informed neural framework for damage analysis and segmentation from a single RGB image. Our contributions are threefold: i) a hierarchical deep learning framework that integrates object-shape estimation with surface damage segmentation, using inferred missing regions to guide predictions and intermediate outputs for traceability; ii) a curated dataset of movable CH artifacts with pixel-level annotations distinguishing surface damage from missing parts; and iii) an empirical evaluation against state-of-the-art methods, demonstrating \emph{DASCH-Net}'s effectiveness in low-resource CH settings. Results show that conditioning damage prediction on estimated shape constrains it to geometrically plausible regions and enables identifying missing parts beyond directly visible evidence. \emph{\href{https://github.com/THMIRALab/DASCH-Net}{Source code}}.
\end{abstract}


\maketitle

\section{Introduction}
Cultural Heritage (CH) artifacts are frequently encountered in a damaged or degraded state, whether due to natural processes over time~\cite{sesana2021climate} or anthropogenic activities (e.g., theft~\cite{waxman2025loot}, war~\cite{ivashko2024destruction}, human negligence~\cite{simmons2024things}). Museums and digital repositories typically document these artifacts with 2D photographs only after damage occurred~\cite {richards2017twenty}. As such, for curators, archaeologists, and CH experts, accurately identifying and localizing damaged regions is essential to support digital reconstruction~\cite{girbacia2024analysis}, inform conservation strategies~\cite{zhang2024archgpt}, facilitate object identification and comparison in trafficking and illicit-trade investigations~\cite{enigma2025}, and document conservation interventions~\cite{korro2025new}.

\begin{figure}[t!]
    \centering
    \begin{subfigure}[b]{0.3\textwidth}
        \centering
        \includegraphics[height=2.8cm]{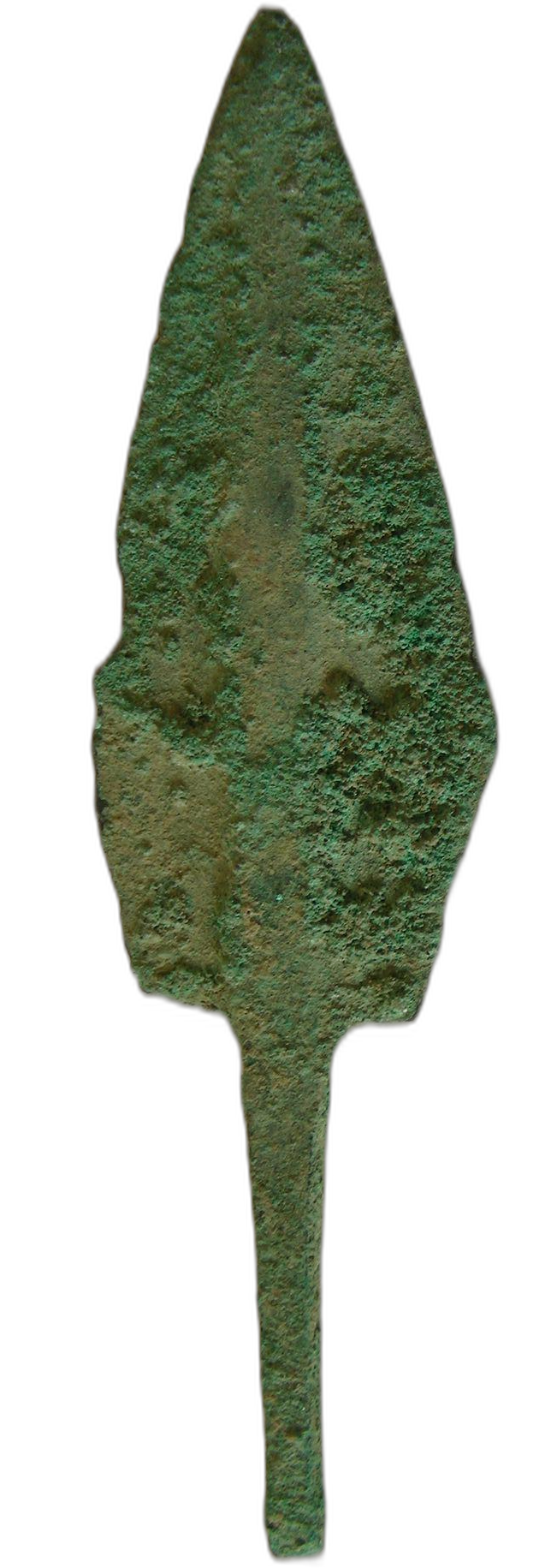}
        \caption{Arrowhead with corrosion~\cite{damaged_arrowhead}.}
        \label{subFigure:damaged-arrowhead}
    \end{subfigure}
    \hspace{0.02\textwidth}
    \begin{subfigure}[b]{0.3\textwidth}
        \centering
        \includegraphics[height=2.8cm]{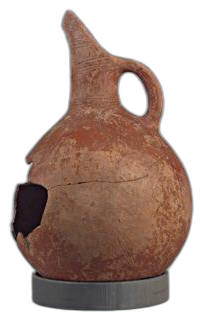}
        \caption{Vase with missing part~\cite{damaged_vase}.}
        \label{subFigure:damaged-vase}
    \end{subfigure}
    \hspace{0.02\textwidth}
    \begin{subfigure}[b]{0.3\textwidth}
        \centering
        \includegraphics[height=2.8cm]{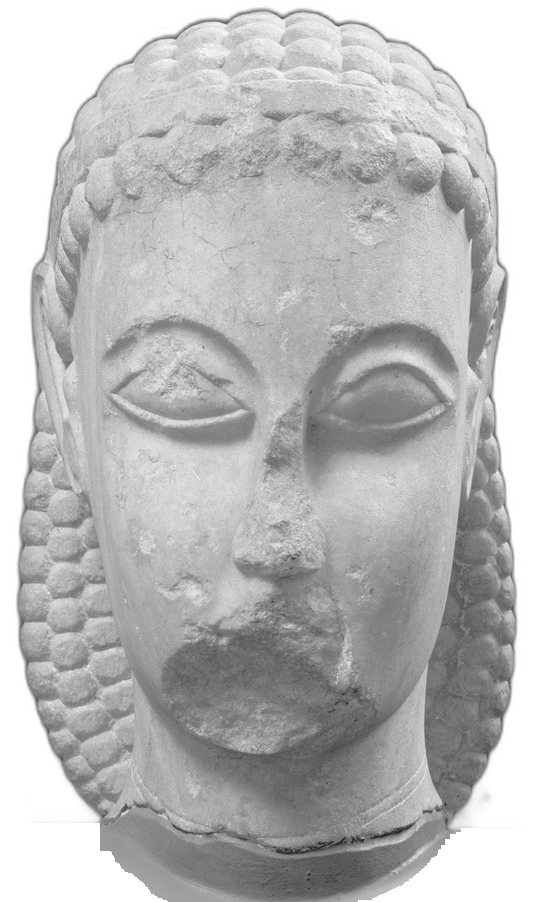}
        \caption{Bust with facial damage~\cite{damaged_bust}.}
        \label{subFigure:damaged-bust}
    \end{subfigure}

    \vspace{0.2cm}

    \begin{subfigure}[b]{0.3\textwidth}
        \centering
        \includegraphics[height=2.8cm]{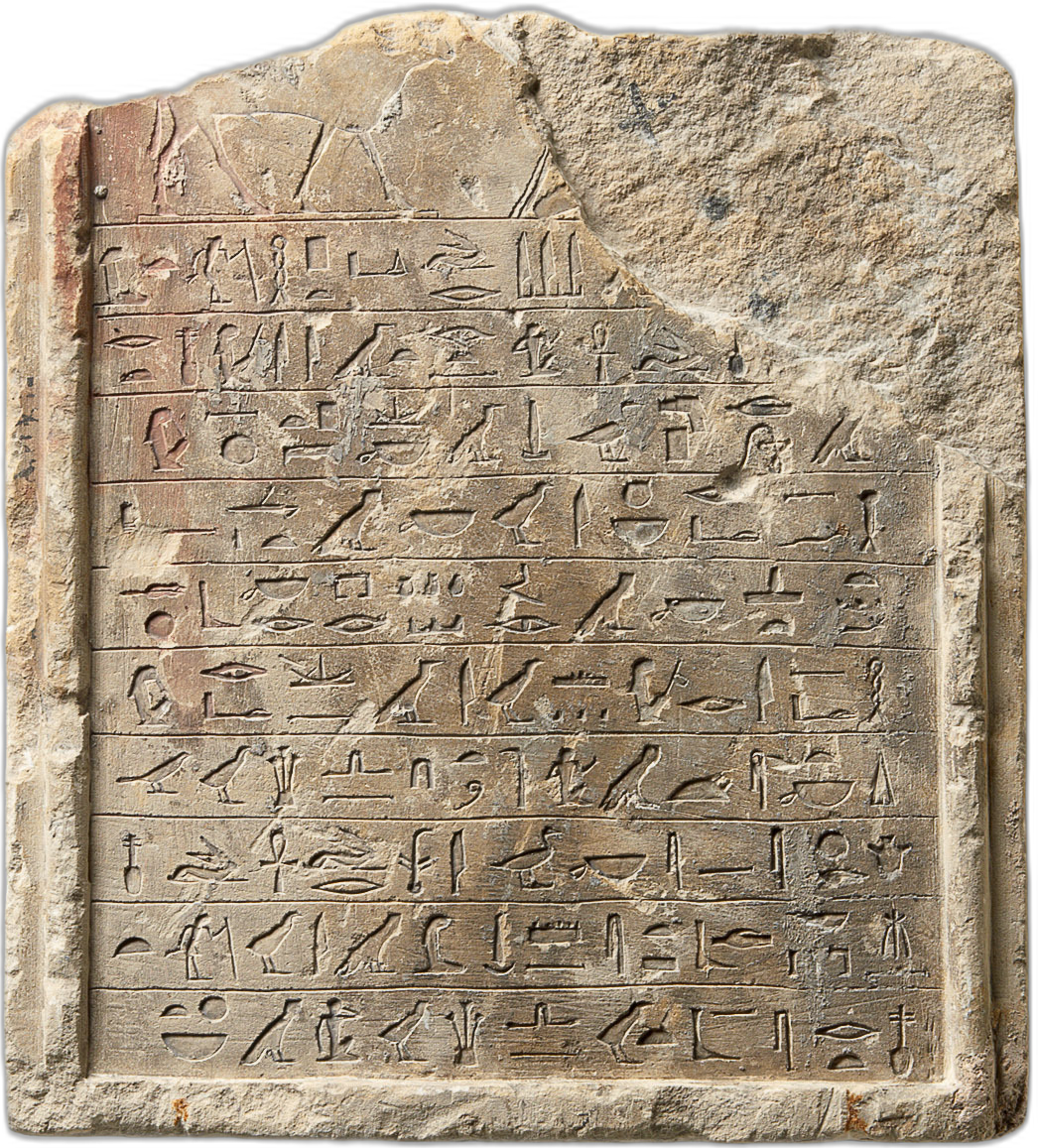}
        \caption{Stele with surface erosion~\cite{damaged_stele}.}
        \label{subFigure:damaged-stele}
    \end{subfigure}
    \hspace{0.02\textwidth}
    \begin{subfigure}[b]{0.3\textwidth}
        \centering
        \includegraphics[height=2.8cm]{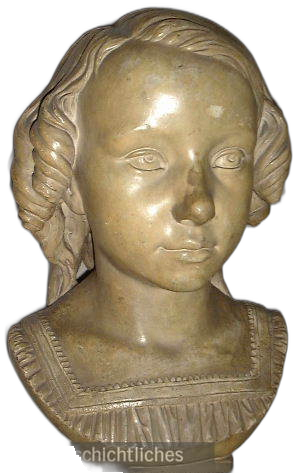}
        \caption{Reflective bust~\cite{statue_reflective}.}
        \label{subFigure:reflective-bust}
    \end{subfigure}
    \begin{subfigure}[b]{0.3\textwidth}
        \centering
        \includegraphics[height=2.8cm]{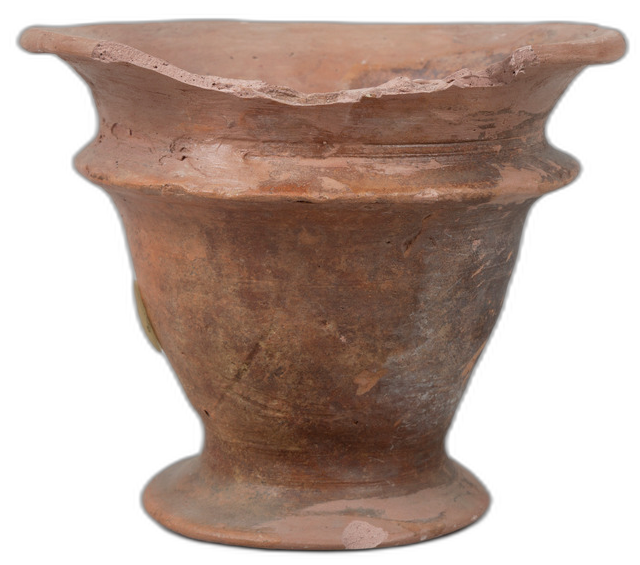}
        \caption{Vase with damaged front~\cite{damaged_vase_front}.}
        \label{subFigure:damage-vase-front}
    \end{subfigure}

    \Description{Six images illustrating typical CH artifacts and damage scenarios: (a) a vase with missing fragments, (b) a bust with facial damage affecting the nose and mouth, (c) a bust with a reflective region in the 2D image, (d) a stone stele with surface erosion and loss of detail, (e) a bronze arrowhead showing corrosion, and (f) a hollow vase with a missing frontal section exposing the interior surface.}

    \caption{Examples of CH artifacts and damage types, including missing structural parts, facial losses in sculptural works, surface erosion of stone inscriptions, corrosion in metal artifacts, and missing frontal sections in hollow objects.}
    \label{fig:examples-damage}
    \vspace{-6pt}
\end{figure}

Motivated by these practical needs, we focus on two significant categories of physical damage observed in CH artifacts. The first is \emph{surface degradation}~\cite{mishra2024artificial}, which affects the object's visible appearance and includes wear, erosion, and cracking (Figures~\ref{subFigure:damaged-arrowhead}, ~\ref{subFigure:damaged-bust}, and~\ref{subFigure:damaged-stele}). The second is \emph{structural damage}~\cite{kantaros2023three}, where missing fragments or broken sections alter the object’s overall geometry and silhouette (Figures~\ref{subFigure:damaged-arrowhead}, ~\ref{subFigure:damaged-vase}, and~\ref{subFigure:damage-vase-front}). While these two forms of damage can occur together, they pose different challenges for visual analysis. Surface damage is directly observable in the image, as its characteristics are visible through texture and material cues. In contrast, structural damage involves missing parts that are absent from the image and must be inferred to estimate the object’s original shape.

This distinction introduces several challenges for computational CH damage analysis. Damage patterns are irregular, multi-scale, and highly dependent on object type, material, surface treatment, and acquisition conditions, ranging from small missing parts to large structural losses with complex boundaries (Figures~\ref{subFigure:damaged-arrowhead} and~\ref{subFigure:damaged-vase}). Reflective surfaces, heterogeneous textures, and field photography further complicate separating damage from visual artifacts (Figures~\ref{subFigure:damaged-bust},~\ref{subFigure:damaged-stele}, and~\ref{subFigure:reflective-bust}). In addition, single-view RGB images introduce depth ambiguity, especially for hollow objects such as vases, where exposed inner surfaces may visually blend with exterior regions (Figure~\ref{subFigure:damage-vase-front}).

As a result, CH damage analysis relies on expert inspection and manual documentation~\cite{karimi2024deep}. Although effective, this process is time-consuming, difficult to scale, and contributes to the scarcity of pixel-level damage annotations required for data-driven methods~\cite{mishra2024artificial}. Recent computational approaches have explored automated damage detection in constrained settings, including paintings, architectural surfaces, built-heritage damage segmentation, disaster damage, and 3D representations~\cite{mishra2024artificial,yue2025automatic,yue2026dual,sanchez2023detection}. These works advance deep learning for CH damage assessment, but they focus on specific object types or damage manifestations, limiting their applicability to diverse movable artifacts.

At the same time, advances in Computer Vision (CV), particularly Vision Foundation Models (VFMs) trained on large-scale visual data~\cite{awais2025foundation}, provide transferable representations that can be adapted to downstream tasks~\cite{oquab2023dinov2,ravi2025sam}. However, their direct use for CH damage analysis remains challenging because damage is domain-specific, sparsely represented in generic training data, and often requires both pixel-level localization and reasoning about missing geometry. Moreover, binary ``damage vs. no-damage'' predictions are insufficient for CH applications, where conservation and curatorial workflows require spatial evidence showing where damage is located, how it relates to the artifact's visible and inferred structure, and which visual or structural cues support the model's decisions~\cite{diaz2020accessible}. 

To address these challenges, we propose \emph{DASCH-Net} (Damage Analysis and Segmentation in Cultural Heritage; Section~\ref{sec:dasch-net}), a hierarchical shape-informed framework for single-view CH damage analysis. \emph{DASCH-Net}'s main contribution lies in formulating damage segmentation as a shape-conditioned problem in which damage prediction is guided by an estimate of the object's plausible intact structure. The framework consists of two interacting components: structural shape estimation and surface degradation detection. The shape component combines a learned shape prior with a visible-silhouette estimate from SAM2~\cite{ravi2025sam} to infer the object's plausible complete shape and identify candidate missing regions. The damage component leverages semantic representations from DINOv2~\cite{oquab2023dinov2}, adapted through lightweight modules~\cite{houlsby2019parameter}, and uses the estimated shape information to guide damage prediction. This hierarchical design enables the model to distinguish visible surface degradation from missing structural parts, constrain predictions to geometrically plausible regions, and support process-level transparency through inspectable intermediate outputs.~\footnote{In this work, process-level transparency refers to the traceability of the prediction process through semantically meaningful intermediate outputs. Specifically, \emph{DASCH-Net} exposes the visible silhouette, estimated complete shape, inferred missing-region cues, and final damage mask, allowing users to inspect how structural evidence guides damage localization.}

We also introduce a curated dataset of damaged CH artifacts (Section~\ref{sec:dataset}) covering diverse object types, materials, cultural contexts, and historical periods. For the damaged objects, we provide pixel-level annotations of both surface damage and missing parts. Through comparisons with state-of-the-art segmentation methods, quantitative experiments, qualitative analysis, interpretability evaluation, and ablation studies, we show that this integration provides a strong reference framework for low-resource CH damage segmentation, particularly when damage localization benefits from explicit shape-aware reasoning (Sections~\ref{sec:experiments} and~\ref{sec:results-discussion}).

In summary, this paper makes the following contributions:
\begin{itemize}
    \item A hierarchical shape-informed framework for single-view CH damage analysis that integrates plausible object-shape estimation with surface damage segmentation, using inferred missing-region cues to guide damage prediction and support process-level transparency through intermediate outputs.
    \item A curated dataset of damaged CH artifacts with pixel-level annotations of surface damage and missing parts, covering diverse object types, materials, cultures, and historical periods.
    \item Empirical experiments against SoA methods, supported by quantitative results, qualitative analysis, interpretability, and ablation studies, demonstrating \emph{DASCH-Net}'s effectiveness in low-resource CH damage segmentation.
\end{itemize}

\section{Literature Review}
\label{sec:literature-review}
We first examine how damage is defined and analyzed in CH practice and how these definitions influence available training data, annotation protocols, and evaluation (Section~\ref{subSec:lit-review-damaged-ch}). We then discuss learning paradigms for semantic segmentation under limited supervision (Section~\ref{subSec:lit-review-damage-scarcity}). Finally, we review interpretability methods, motivated by the need to diagnose model behavior and support expert trust (Section~\ref{subSec:lit-review-interpretability}).

\subsection{Damage Analysis and Datasets in Cultural Heritage}
\label{subSec:lit-review-damaged-ch}
In the CH literature, damage broadly refers to physical, chemical, or biological alterations that compromise an artifact’s integrity. Damage is inherently heterogeneous and context-dependent, shaped by material properties, environmental exposure, biological activity, and historical processes~\cite{stoean2023computational}. For organic materials (e.g., wood, textiles, papyrus, parchment), biological deterioration may cause discoloration, staining, fiber weakening, structural degradation, and material loss~\cite{gadd2024fungal,paolino2024greener}. As a result, damage manifestations vary widely across artifact types, leading existing computational approaches to focus on specific materials, damage patterns, or acquisition conditions~\cite{ranieri2025automated,zhang2024application}.

From a computational perspective, this diversity is reflected in a clear distinction between how surface-level degradation, which manifests as visible texture, color, or material alterations and can often be addressed using appearance cues~\cite{garcia2025artinsight,yue2026dual}, and structural loss are addressed, where missing regions alter object geometry and must be inferred rather than detected~\cite{villegas2023matchmakernet,jaramillo2024cultural}. Most existing methods specialize in one of these aspects, resulting in architectural limitations: appearance-driven approaches struggle with missing regions and boundary coherence, while structure-oriented methods may overlook subtle surface damage~\cite{xiao2021amodal,cheng2026comprehensive}. When surface degradation and structural loss co-occur, treating them independently can reduce accuracy and interpretability~\cite{yu2025cross}. In response, recent studies across different fields increasingly point toward pixel-level semantic segmentation augmented with structure-aware reasoning as a promising direction for robust damage analysis~\cite{gao2023coarse,wang2025bimodal}, highlighting the need for unified frameworks that explicitly integrate surface degradation and structural loss within a single damage identification pipeline, as adopted in this work by ~\emph{DASCH-Net}.

These methodological choices are reflected in available datasets. Existing resources provide pixel-level damage annotations for paintings and murals~\cite{xu2024comprehensive,stoean2023computational}, microscopic metal surfaces~\cite{garcia2025artinsight}, and heritage walls and buildings~\cite{yue2025automatic,yue2026dual}. Outside CH, datasets address structural damage via cracks, fragment reconstruction, or 3D representations~\cite{benz2024omnicrack30k,sellan2022breaking}. However, these resources rarely provide unified supervision for both surface degradation and structural loss within a single pixel-level framework. Addressing this gap, our work introduces a curated dataset of movable CH artifacts with explicit annotations for both damage types (Section~\ref{sec:dataset}).

\subsection{Surface Degradation and Shape Loss Under Annotation Scarcity}
\label{subSec:lit-review-damage-scarcity}

\textbf{Surface damage under limited supervision.} Recent advances in semantic segmentation in low-resource settings have shown that accurate pixel-level localization of defects is feasible even when dense annotations are scarce, as demonstrated in medical image analysis, industrial inspection, and infrastructure monitoring~\cite{jiang2024ph,kim2024few,xiang2023semi}. Prior work explores semi- and weakly supervised segmentation~\cite{yang2024unimatch}, anomaly detection formulations~\cite{bergmann2021mvtec}, and transfer learning approaches~\cite{zhang2024improving}. These strategies are relevant to movable CH artifacts because RGB inspection of bridges, pavements, and industrial surfaces similarly requires learning sparse damage cues under variable appearance and limited supervision. However, while effective for appearance-based defects, these methods typically conceptualize degradations as localized surface irregularities and do not explicitly model object-level structure or missing-part phenomena. When transferred to CH damage analysis, this assumption introduces intrinsic limitations. Surface degradation and structural loss frequently co-occur, and accurate localization depends on understanding object extent and expected geometry. Therefore, appearance-driven segmentation methods struggle with incomplete silhouettes. Motivated by these limitations, \emph{DASCH-Net} integrates surface-level segmentation with shape reasoning, enabling damage predictions to be informed by estimated object shape.

\textbf{Adapting vision foundation models.} Vision Foundation Models (VFMs), pretrained on large-scale visual data, provide transferable representations that generalize well under limited supervision~\cite{awais2025foundation}. Representative VFMs include segmentation-oriented models (e.g., SAM2~\cite{ravi2025sam}) and self-supervised encoders (e.g., DINOv2~\cite{oquab2023dinov2}). To adapt these models efficiently, recent work emphasizes adapter-based parameter-efficient fine-tuning~\cite{houlsby2019parameter}. \emph{DASCH-Net} builds on these advances by jointly leveraging two VFMs. Specifically, DINOv2 for semantic surface representation and SAM2 for structure-aware object extent estimation. We integrate and adapt both models using lightweight adapters.

\textbf{Shape priors for missing parts.} Related work on amodal and image completion addresses reasoning about invisible object parts~\cite{xiao2021amodal,ao2023image}. Existing approaches span generative models (e.g., GAN~\cite{ao2023image}, diffusion-based frameworks~\cite{tran2024amodal}), transformer-based methods~\cite{gao2023coarse}, and more recent foundation-model extensions that enable zero-shot amodal segmentation~\cite{liu2025towards}. While conceptually relevant, these methods typically assume intact objects with synthetic or transient occlusions, whereas CH scenarios involve irreversible material loss. For hollow or fragmented artifacts, missing parts alter object boundaries and global geometry, limiting the applicability of generic amodal pipelines. To address this, \emph{DASCH-Net} incorporates structural priors through a VAE-based shape model learned from intact artifacts and a SAM2-based silhouette estimator. These priors guide missing-part inference and inform both shape estimation and surface damage localization.

\subsection{Interpretability for Damage in CH Artifacts}
\label{subSec:lit-review-interpretability}
Interpretability is essential in CH applications, where automated predictions must support expert validation and conservation decision-making~\cite{diaz2020accessible,matrone2023explaining}. Prior work emphasizes that opaque damage predictions can hinder trust, particularly in heterogeneous and data-scarce settings~\cite{pansoni2023artificial}. For dense prediction tasks, attribution-based methods widely link model outputs to spatial image evidence~\cite{gipivskis2024explainable}. Among these, Grad-CAM and its variants~\cite{selvaraju2017grad} are particularly well suited to segmentation models because they use intermediate feature representations. Given that surface degradation and missing-part damage rely on different visual cues, this work adopts a \emph{decoupled interpretability strategy}. We apply Grad-CAM separately to the shape estimation and surface-damage components of \emph{DASCH-Net}, enabling fine-grained analysis of how structural priors influence object-extent reasoning and how appearance cues drive localized damage responses. 

\section{Background}
\label{sec:background}
This section introduces the technical concepts upon which this
work is built. First, we introduce VAEs as a mechanism for learning
global shape priors that support inference for artifacts with
missing regions (Section~\ref{subSec:back-vae}). We then review
VFMs as feature extractors that provide transferable spatial and
semantic representations (Section~\ref{subSec:back-vfm}), focusing
on SAM2-Hiera and DINOv2. Third, we describe lightweight
adaptation strategies based on adapter modules, which enable
task-specific specialization of VFMs with a small number of
trainable parameters (Section~\ref{subSec:back-adapter}). Finally, we discuss
process-level transparency and gradient-based attribution
methods that support qualitative inspection of model predictions
(Section~\ref{subSec:back-transparency}).

\begin{figure}[t!]
    \centering
    \includegraphics[width=0.80\textwidth]{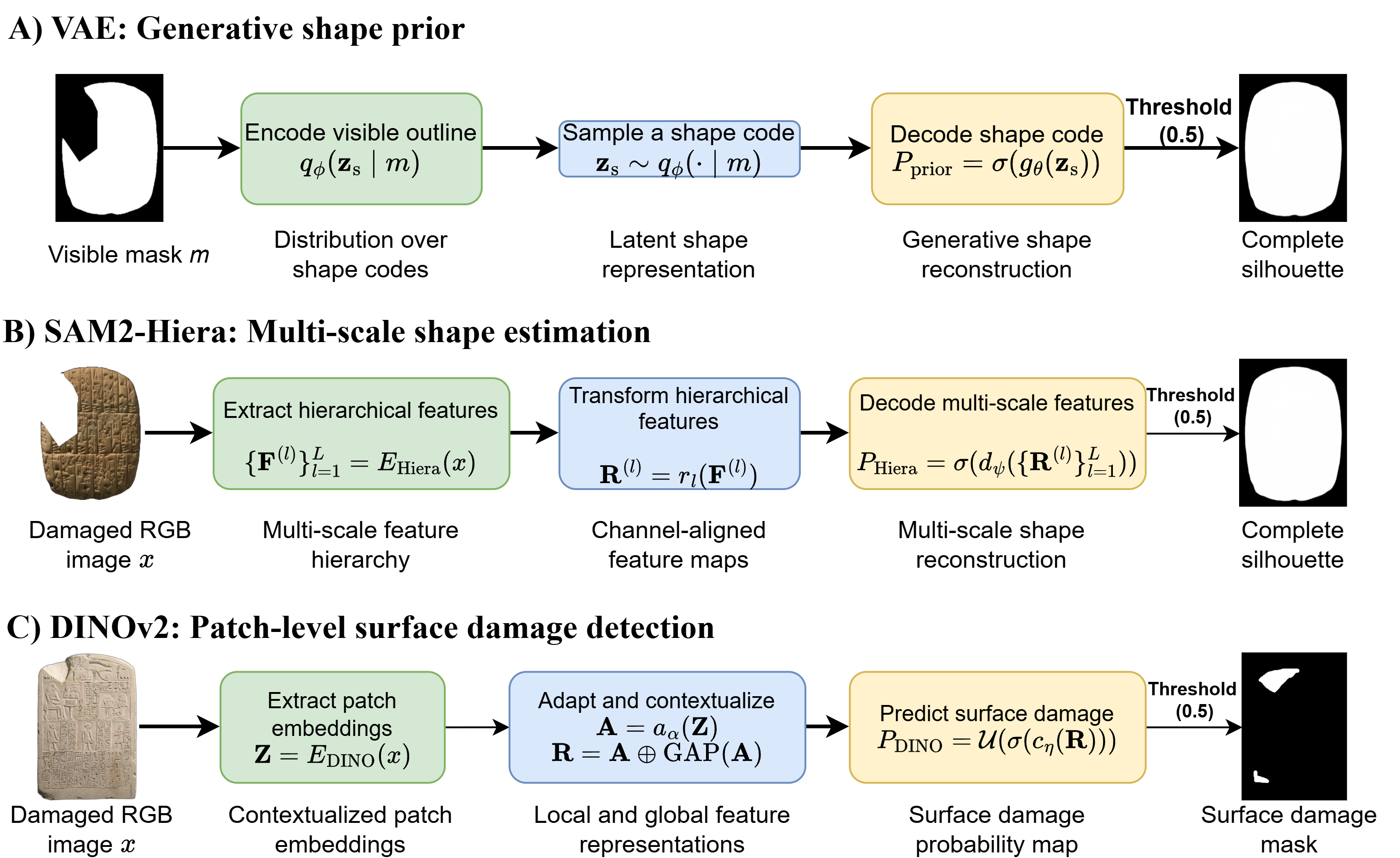}

    \Description{Three horizontal diagrams illustrating the
    prediction mechanisms used in DASCH-Net. Panel A transforms
    a visible object mask into an estimated complete silhouette
    through variational encoding, latent sampling, and decoding.
    Panel B transforms an RGB image into a complete silhouette
    through hierarchical feature extraction, transformation,
    and multi-scale decoding. Panel C transforms an RGB image
    into a surface damage mask through patch-level feature
    extraction, adaptation, contextualization, and damage
    prediction. Each diagram includes the mathematical
    formulation of its successive operations.}

    \caption{Three predictions used by \emph{DASCH-Net}. A) The VAE estimates a plausible complete silhouette from the visible mask. B) The SAM2-Hiera branch estimates a complete silhouette from the RGB photograph. C) The DINOv2 branch estimates surface damage from local patches and their image context.} 

    \label{fig:background-transformations}
\end{figure}

To illustrate the relevance of these concepts to CH conservation,
Figure~\ref{fig:background-transformations} presents the three base
prediction mechanisms in \emph{DASCH-Net}. From the visible silhouette, the VAE hypothesizes how the original outline may have extended beyond the damaged boundaries (A). SAM2-Hiera complements this geometric prior by examining the surviving boundaries and broader structure of the artifact in the RGB data (B). DINOv2 focuses on surface appearance to localize visible damage (C). Their combination supports the assessment of an artifact's state of conservation from a \emph{single} photograph. Their applicability to CH
damage analysis is examined in Section~\ref{sec:results-discussion}.

\subsection{Variational Autoencoder (VAE) for Shape Priors}
\label{subSec:back-vae}

We use Variational Autoencoders (VAEs)~\cite{kingma2019introduction} to learn shape priors from intact object silhouettes, constraining shape inference toward geometrically plausible configurations. During training, we remove regions from intact silhouettes and use the original masks as reconstruction targets. In this way, VAE learns relationships between partial observations and
plausible complete shapes. Figure~\ref{fig:background-transformations}A illustrates how we generate a shape prior at inference. Given the visible mask $m \in \{0,1\}^{H \times W}$ ($H$ and $W$ represent
the image height and width) of an artifact,
this process unfolds as follows:
\begin{equation}
\label{eq:background-shape-vae}
q_{\phi}(\mathbf{z}_{\mathrm{s}} \mid m),
\qquad
\mathbf{z}_{\mathrm{s}} \sim q_{\phi}(\cdot \mid m),
\qquad
P_{\mathrm{prior}} = \sigma(g_{\theta}(\mathbf{z}_{\mathrm{s}})).
\end{equation}
The encoder $q_{\phi}$, with learned weights $\phi$, estimates a probability distribution over latent shape
representations $\mathbf{z}_{\mathrm{s}}$ conditioned on the
visible mask $m$, from which a shape
code $\mathbf{z}_{\mathrm{s}}$ is sampled, providing a compact representation of the artifact's shape. The decoder $g_{\theta}$, with learned weights $\theta$, maps this code to reconstruction
logits. The sigmoid function $\sigma$ converts these logits into
a probability map \(P_{\mathrm{prior}}\in[0,1]^{H\times W}\), whose values estimate the probability of each pixel belonging to the complete silhouette.
Thresholding this map at $0.5$ yields the estimated complete
silhouette.

\subsection{Vision Foundation Model (VFM) for Spatial and Semantic Representation}
\label{subSec:back-vfm}
VFMs~\cite{awais2025foundation} learn transferable spatial
and semantic representations from large and diverse image
collections. These representations are relevant
to CH damage analysis because artifacts exhibit considerable
variability in shape, material, surface appearance, and damage
patterns, while annotated damaged samples remain scarce.
We leverage SAM2-Hiera to capture structural information
for complete shape estimation and DINOv2 to capture visual
appearance cues for surface damage identification. We describe these VFMs in detail in the following sections. 

\subsubsection{SAM2 for Shape Estimation}
\label{subSubSec:back-sam2}
Segment Anything Model 2 (SAM2)~\cite{ravi2025sam} is a promptable segmentation model that produces object masks. Its image encoder, Hiera, is a hierarchical vision
transformer that progressively reduces spatial resolution while
increasing feature dimensionality, capturing fine-grained
boundaries and broader object structure. We leverage these
multi-scale representations to estimate complete object
silhouettes from RGB images of damaged
artifacts (Figure~\ref{fig:background-transformations}B). Given an input image $x \in \mathbb{R}^{H \times W \times 3}$ ($H$ and $W$ denote its spatial dimensions and $3$ its
color channels), the shape estimation process is formulated as:
\begin{equation}
\label{eq:sam2_representation}
\{\mathbf{F}^{(l)}\}_{l=1}^{L} = E_{\mathrm{Hiera}}(x),
\quad
\mathbf{R}^{(l)} = r_l(\mathbf{F}^{(l)}),
\quad
P_{\mathrm{Hiera}} =
\sigma(d_{\psi}(\{\mathbf{R}^{(l)}\}_{l=1}^{L})).
\end{equation}
The encoder $E_{\mathrm{Hiera}}$ extracts a hierarchy of $L$
feature maps, where $l$ indexes the hierarchy level and
$\mathbf{F}^{(l)} \in \mathbb{R}^{H_l \times W_l \times C_l}$
denotes a feature map with spatial dimensions $H_l$ and $W_l$,
and $C_l$ denotes the number of learned feature channels per
spatial location. Each receptive field block $r_l$ transforms
the corresponding feature map into a representation
$\mathbf{R}^{(l)}$ with a common channel dimensionality. The decoder $d_{\psi}$, with learned parameters
$\psi$, progressively combines these representations from coarse
to fine spatial resolutions. The sigmoid function $\sigma$
converts the resulting logits into a probability map
$P_{\mathrm{Hiera}} \in [0,1]^{H \times W}$. Thresholding this map at $0.5$ yields the estimated complete
silhouette.

\subsubsection{DINOv2 for Surface Damage}
\label{subSubSec:back-dinov2}
DINOv2~\cite{oquab2023dinov2} is a self-supervised vision
foundation model that extracts dense, patch-level semantic
representations from RGB images. We leverage
these representations to identify surface damage in the artifact (Figure~\ref{fig:background-transformations}C). Given an RGB
image $x \in \mathbb{R}^{H \times W \times 3}$, the surface
damage estimation process is formulated as:
\begin{equation}
\label{eq:dino_tokens}
\mathbf{Z} = E_{\mathrm{DINO}}(x),
\quad
\mathbf{A} = a_{\alpha}(\mathbf{Z}),
\quad
\mathbf{R} = \mathbf{A} \oplus \operatorname{GAP}(\mathbf{A}),
\quad
P_{\mathrm{DINO}} = \mathcal{U}(\sigma(c_{\eta}(\mathbf{R}))).
\end{equation}
The frozen encoder $E_{\mathrm{DINO}}$ partitions the image
into $N = HW/P^2$ non-overlapping patches of size $P \times P$
and extracts a sequence of embeddings
$\mathbf{Z} = (\mathbf{z}_1,\ldots,\mathbf{z}_N)$, where
$\mathbf{z}_i \in \mathbb{R}^{D}$ denotes the $D$-dimensional
representation of patch $i$. These embeddings are rearranged
into a spatial grid of dimensions $H_p = H/P$ and $W_p = W/P$
and transformed by the adapter $a_{\alpha}$, providing an adapted
feature map $\mathbf{A} \in \mathbb{R}^{H_p \times W_p \times D}$. Global average pooling ($\operatorname{GAP}$) aggregates
the adapted features into a global descriptor, which is
broadcast across spatial locations and concatenated
channel-wise ($\oplus$) with $\mathbf{A}$, resulting in the contextualized feature map $\mathbf{R} \in \mathbb{R}^{H_p \times W_p \times 2D}$.
The convolutional classifier $c_{\eta}$, with learned
parameters $\eta$, integrates information from neighboring
patches to produce patch-level damage logits. The sigmoid
function $\sigma$ converts these logits into probabilities,
which are upsampled through bilinear interpolation
$\mathcal{U}$ to obtain the surface damage probability map
$P_{\mathrm{DINO}} \in [0,1]^{H \times W}$. We can then threshold this map at $0.5$ to obtain the estimated surface damage mask.

\subsection{Adapter for Lightweight Fine-Tuning}
\label{subSec:back-adapter}
While VFMs provide strong generic representations, adapting them
to CH damage analysis requires task-specific specialization.
Adapters introduce lightweight bottleneck modules that refine frozen
backbone representations with a small number of additional parameters.
Houlsby et al.~\cite{houlsby2019parameter} introduce adapters as
feed-forward modules composed of a projection to a low-dimensional
bottleneck, a non-linear activation, and a projection back to the
original feature dimension, combined residually with the backbone
representation. Given an intermediate feature representation
$\mathbf{h} \in \mathbb{R}^{D}$ and a bottleneck dimension
$d \ll D$, the bottleneck component is formulated as:
\begin{equation}
\label{eq:adapter_bottleneck}
a_{\alpha}(\mathbf{h}) =
\mathbf{W}_{\mathrm{up}}
\rho(\mathbf{W}_{\mathrm{down}}\mathbf{h}),
\end{equation}
where $\mathbf{W}_{\mathrm{down}} \in \mathbb{R}^{d \times D}$
and $\mathbf{W}_{\mathrm{up}} \in \mathbb{R}^{D \times d}$
denote the learnable down- and up-projection matrices,
respectively, and $\rho$ denotes a non-linear activation function.
$\alpha$ represents all trainable parameters
of the adapter. Optimizing $\alpha$ enables task-specific
adaptation of the feature representations while keeping the
pretrained backbone parameters frozen. In this work, adapters allow us to specialize the pretrained representations of SAM2-Hiera and DINOv2 for CH artifacts.

\subsection{Process-Level Transparency and Visual Diagnostics}
\label{subSec:back-transparency}
The integration of the previously introduced components in
\emph{DASCH-Net} supports process-level transparency by making
intermediate predictions accessible for inspection. Specifically,
the Shape VAE and SAM2-Hiera provide complementary estimates
of the complete silhouette, while DINOv2 identifies surface
damage from visual appearance. These outputs allow us to
inspect how geometric and appearance-based evidence is
integrated into the final damage prediction. To further
investigate the spatial evidence associated with individual
predictions, we employ Gradient-weighted Class Activation
Mapping (Grad-CAM)~\cite{selvaraju2017grad} as a complementary
post-hoc diagnostic tool. Formally, given a differentiable
scalar prediction score $y$, which in segmentation tasks can
be obtained by aggregating predictions over a spatial region
of interest, and a set of intermediate activation maps
$\mathbf{B} = (B^k)_{k=1}^{K}$, Grad-CAM computes:
\begin{equation}
\label{eq:gradcam}
w_k = \frac{1}{uv}
\sum_{i=1}^{u}\sum_{j=1}^{v}
\frac{\partial y}{\partial B^k_{ij}},
\quad
L_{\mathrm{GradCAM}} =
\mathrm{ReLU}\left(\sum_{k=1}^{K} w_k B^k\right).
\end{equation}
Here, $K$ denotes the number of feature channels,
$B^k \in \mathbb{R}^{u \times v}$ is the activation map of
channel $k$ with spatial dimensions $u$ and $v$, and
$B^k_{ij}$ is its activation at location $(i,j)$.
The importance weight $w_k$ quantifies the relevance of
channel $k$ to the prediction score $y$ by averaging its
gradients across all spatial locations. The weighted activation
maps are combined and passed through the ReLU function,
yielding a spatial attribution map
$L_{\mathrm{GradCAM}} \in \mathbb{R}_{\geq 0}^{u \times v}$
that highlights regions contributing positively to the prediction. In practice, these maps allow us to assess whether shape estimation and
damage identification rely on geometrically and semantically
relevant regions of the artifact
(see Figures~\ref{fig:results-shape-prior-interpretability}
and~\ref{fig:results-surface-damage-interpretability}).

\section{DASCH-Net: A Shape-Conditioned Framework for Damage Analysis and Segmentation in CH}
\label{sec:dasch-net}
\begin{figure}[t!]
    \centering
    \includegraphics[width=0.8\textwidth]{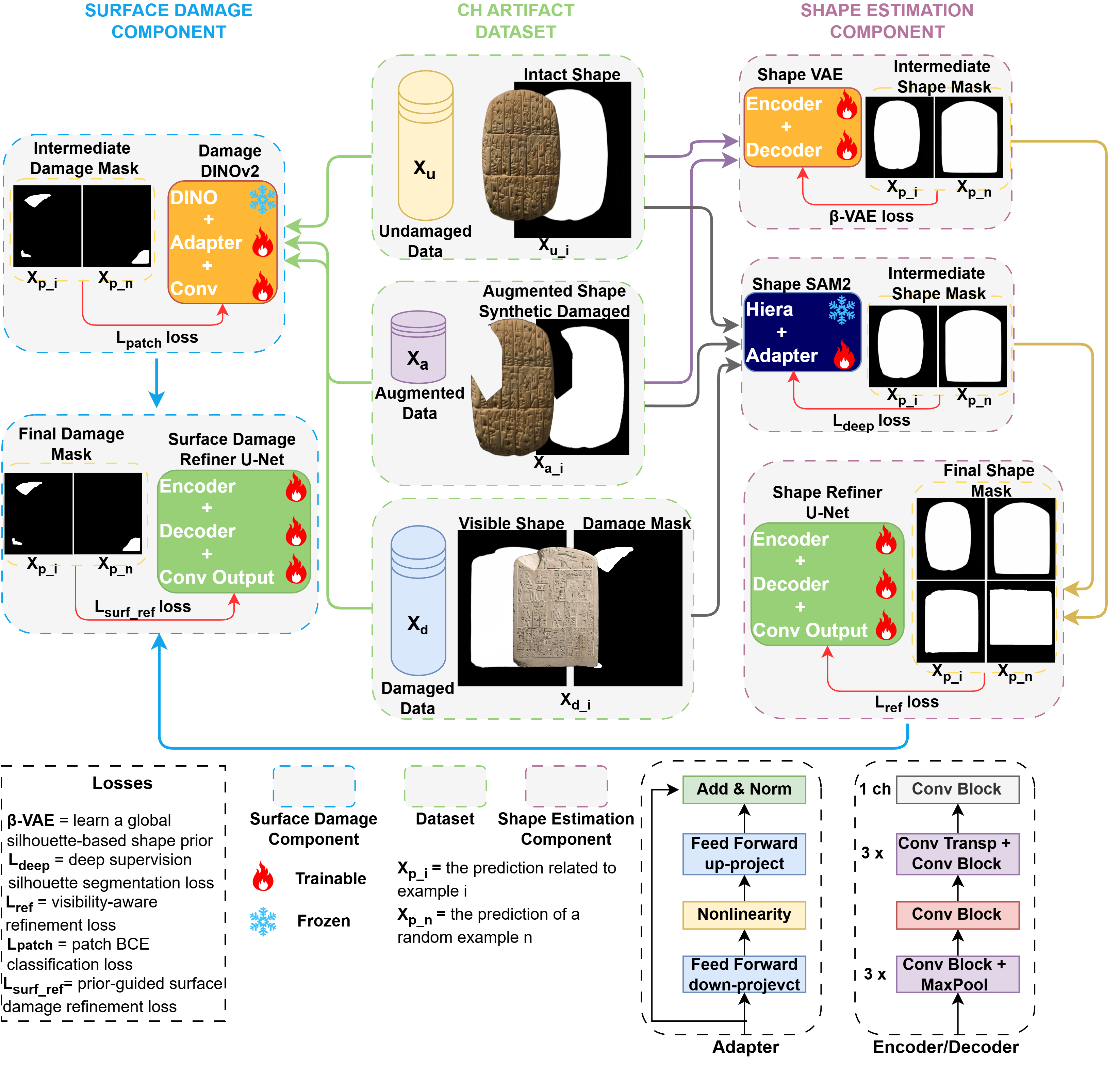}
    \Description{Diagram of the DASCH-Net framework. DASCH-Net includes a silhouette-based shape prior learned with a variational autoencoder, a SAM2-Hiera network for shape estimation and refinement, and a DINOv2-based module for patch-level surface damage detection. Shape and damage predictions are progressively refined using U-Net–style networks, and the final surface damage estimation incorporates the refined shape prior and visibility information.}
    
    \caption{The DASCH-Net framework. The diagram should be read by following the arrows from the central CH artifact dataset toward the two processing components. The dataset provides intact, synthetically damaged, and real damaged samples that feed both the shape-estimation and surface-damage components. The framework is organized as a multi-stage pipeline that progressively integrates shape priors and appearance cues. First, a Shape VAE learns a prior over plausible complete silhouettes. A SAM2-Hiera shape-estimation branch and a U-Net refiner then produce the final amodal shape estimate. In parallel, a DINOv2-based branch detects surface damage from appearance cues, and a final refiner combines damage, visibility, and shape information to produce the complete damage mask.}

    \label{fig:dasch-net-framework}
\end{figure}

We introduce \emph{DASCH-Net} (Figure~\ref{fig:dasch-net-framework}), a hierarchical shape-conditioned framework for \textbf{D}amage \textbf{A}nalysis and \textbf{S}egmentation of \textbf{CH} artifacts from a single RGB image. \emph{DASCH-Net} formulates damage analysis as the joint estimation of structural shape and surface degradation. The framework consists of two interacting components: (1) \textbf{structural shape estimation}, which predicts a plausible complete object silhouette accounting for missing parts, and (2) \textbf{surface degradation detection}, which localizes damage while using inferred missing regions to guide pixel-level prediction. Specifically, \emph{DASCH-Net} estimates an amodal object shape and derives missing-region cues from the discrepancy between the visible silhouette and the estimated complete shape. These cues condition damage refinement, enabling the model to combine appearance evidence of surface degradation with geometric evidence of structural loss. This hierarchical design constrains predictions to geometrically plausible regions of objects and supports process-level transparency by making the links between visible damage, inferred missing structure, and final damage localization inspectable through intermediate outputs.

Formally, given an RGB image $x \in \mathbb{R}^{H \times W \times 3}$ and a visible object mask $m \in \{0,1\}^{H \times W}$, the goal is to predict a damage probability map $\hat{y}_d \in [0,1]^{H \times W}$ that captures both surface degradation and missing-part damage. Since missing regions are not directly observable, \emph{DASCH-Net} first estimates a complete object silhouette $\hat{s} \in \{0,1\}^{H \times W}$, which it then uses to guide surface damage refinement. Both components rely on frozen VFMs augmented with adapters and are optimized in multiple stages, reflecting their distinct learning objectives. To mitigate the limited availability of damaged training samples, we augment intact silhouettes with synthetically generated missing parts. Given a complete silhouette $M_{\text{complete}}$, we sample an irregular fracture mask $D \in \{0,1\}^{H \times W}$ that simulates a missing region. The corresponding visible silhouette is then defined as $M_{\text{visible}} = M_{\text{complete}} \wedge \neg D$, followed by standard geometric transforms. This produces diverse missing-part configurations while preserving complete ground truth, and is reused across all stages.

\subsection{Shape Prior Component}
\label{subSec:shape-prior-component}

The shape estimation component outputs a refined amodal silhouette as a probability map $P_{\text{ref}} \in [0,1]^{H \times W}$ and a binary mask $\hat{s} \in \{0,1\}^{H \times W}$. We train this component in three stages:

\textbf{Stage 1: Generative silhouette prior (Shape VAE).} We train a silhouette-based VAE to model the distribution of complete shapes. At inference, conditioned on a visible silhouette, the VAE reconstructs a plausible complete silhouette, yielding a prior hypothesis $P_{\text{prior}}$ that regularizes shape completion toward geometrically consistent solutions (Equation~\ref{eq:background-shape-vae}).

\textbf{Stage 2: VFM-based shape estimation (SAM2-Hiera).} We freeze the SAM2-Hiera encoder and train its adapters and decoder to estimate the complete silhouette from $x$. We supervise predictions at several resolutions, emphasizing pixels present in $M_{\mathrm{complete}}$ and absent from $M_{\mathrm{visible}}$. This branch produces $P_{\mathrm{Hiera}}$ (Equation~\ref{eq:sam2_representation}).

\textbf{Stage 3: Silhouette fusion and refinement.} A U-Net combines $P_{\mathrm{prior}}$, $P_{\mathrm{Hiera}}$, and the visible mask $m$ to produce the refined silhouette probability map $P_{\mathrm{ref}}\in[0,1]^{H\times W}$, enforcing (i) consistency on observed regions and (ii) plausible completion on missing regions. Thresholding $P_{\mathrm{ref}}$ gives the estimated complete silhouette $\hat{s}$.

\subsection{Surface Damage Detection and Shape-Aware Refinement}
\label{subSec:surface-damage-component}
This component follows a two-stage approach that first localizes damage on the artifact's visible surface and then integrates shape information to \emph{jointly} segment surface damage and missing parts. The module outputs a refined probability map $PMap_{\text{ref}} \in [0,1]^{H \times W}$ and a binary mask $\hat{D} \in \{0,1\}^{H \times W}$.

\textbf{Stage 1: Patch-level damage localization (DINOv2).} We train a lightweight patch classifier on top of a frozen DINOv2 encoder to identify visual patterns associated with surface damage, such as cracks and abrasion. Using only surface damage annotations, the classifier learns to estimate damage probabilities for each image patch ($Dam_{\text{patch}}$). Stratified oversampling is applied during training to mitigate class imbalance. Once trained, this module is frozen.

\textbf{Stage 2: Shape-aware pixel refinement.} 
We refine the patch-level predictions by integrating structural information from the shape component. We first upsample $Dam_{\text{patch}}$ to image resolution and derive a candidate missing-region mask $M_{\text{missing}} = \max(0, M_{\text{complete}} - M_{\text{visible}})$ from the estimated complete and visible silhouettes. A refinement U-Net takes as input (i) projected DINOv2 features, (ii) $M_{\text{complete}}$, (iii) $M_{\text{visible}}$, (iv) $M_{\text{missing}}$, and (v) the upsampled patch predictions. Unlike Stage 1, the refiner is trained with annotations of both surface damage and missing parts, using a composite loss that combines BCE and Dice terms with a shape-based penalty to suppress false positives. It outputs a refined probability map $PMap_{\text{ref}} \in [0,1]^{H \times W}$, which is thresholded to obtain the final damage mask $\hat{D} \in \{0,1\}^{H \times W}$.

\section{The Cultural Heritage Damage Dataset}
\label{sec:dataset}
\begin{figure}[t!]
    \centering
    \clipbox{0pt {.5\height} 0pt 0pt}{%
        \includegraphics[width=.30\linewidth]{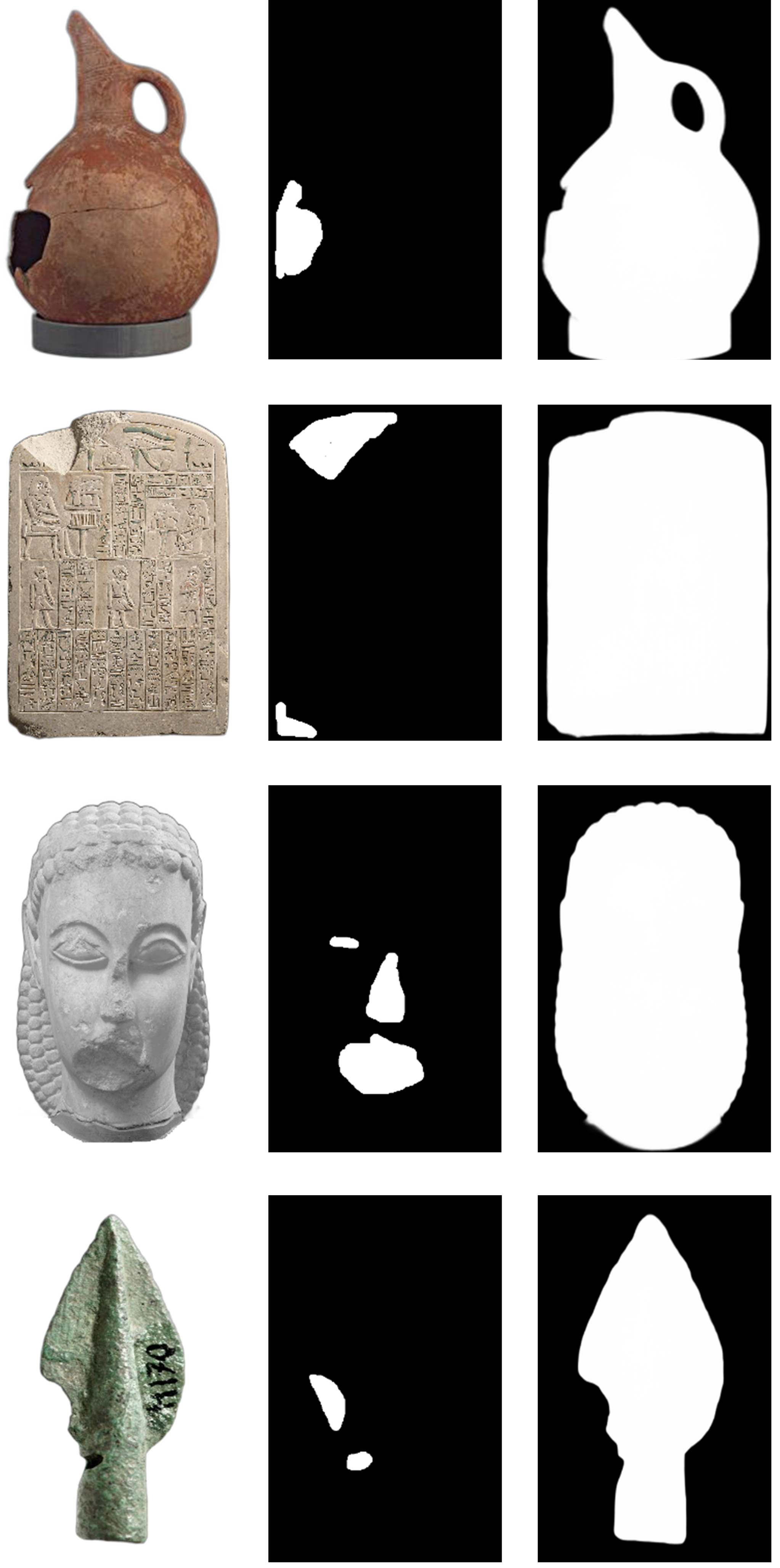}}%
    \hspace{.08\linewidth}%
    \clipbox{0pt 0pt 0pt {.5\height}}{%
        \includegraphics[width=.30\linewidth]{imgs/artifacts/examples_dataset.png}}

    \Description{Four cultural heritage artifacts are arranged in two rows. The top row shows a ceramic vase on the left and a bust on the right. The bottom row shows an inscribed stone tablet on the left and a metal arrowhead on the right. Each example shows the RGB photograph, the binary damage mask, and the binary visible object mask, from left to right. White pixels mark annotated damage in the damage mask and the visible artifact in the object mask, both against a black background.}

    \caption{Examples of four artifact types in the CH damage dataset. For each artifact, the three panels show the original RGB image, the human-annotated ground-truth damage mask, and the visible object mask (from left to right). The visible object mask delineates the artifact's observed silhouette in the original image and may therefore include missing parts when structural damage is present (vase, tablet, arrowhead), whereas it remains intact for purely surface-level degradation (bust).}
    \vspace{-8pt}
    \label{fig:examples-dataset}
\end{figure}

Despite recent advances in damage identification for CH artifacts, datasets of 2D images with pixel-level human annotations remain limited~\cite{wu2024dadnet,mishra2024artificial,xu2024comprehensive}. Existing datasets often focus on specific object categories, cultural contexts, or historical periods, limiting their applicability to digital reconstruction, conservation analysis, and cross-category damage research~\cite{ranieri2025automated}. To address this gap, we introduce a curated dataset of movable CH artifacts from single RGB images, with pixel-level human annotations distinguishing between \emph{surface damage} and \emph{missing parts} (Figure~\ref{fig:examples-dataset}).

The dataset provides three modalities: the RGB image, a pixel-level damage mask indicating surface degradation or missing parts, and a visible object mask obtained with \textit{RemBG}~\footnote{\url{https://pypi.org/project/rembg/}}. This visible object silhouette captures the geometric extent of the artifact as it appears in the image. Consequently, when an artifact exhibits missing parts, the silhouette reflects this incompleteness, whereas in cases of purely surface-level damage, the silhouette remains intact.

The dataset (Table~\ref{tab:ch-dataset-statistics}) comprises 1,746 CH artifacts spanning four object categories, 13 cultural contexts, and diverse historical periods (from the Neolithic period to the 20th century CE), introducing substantial variation in object geometry, materials, and damage patterns. This diversity poses realistic challenges for automated damage analysis, particularly in low-resource settings. Each object is labeled as intact or damaged, and pixel-level damage masks are provided for all damaged artifacts. The dataset is intentionally imbalanced, with 1,027 intact and 719 damaged objects. This imbalance reflects real-world digitization trends, where intact artifacts and certain cultural collections are overrepresented. We preserve this distribution to support evaluation under realistic deployment conditions. 

\begin{table*}[t]
\centering
\small
\begin{tabular}{l l p{4.0cm} c c}
\toprule
\textbf{Object Category} & \textbf{Cultural Context} & \textbf{Period} & \textbf{Quantity} & \textbf{Damaged} \\
\midrule

\multirow{6}{*}{Vase \& Bowl}
& Greek       & 300 BCE -- 1 BCE \& 2000 BCE -- 1700 BCE & 83  & 30 \\
& Peruvian    & Multiple Periods                 & 659 & 372 \\
& Indian      & 200 CE -- 500 CE                     & 22  & 10 \\
& Iranian     & 1800 CE                           & 21  & 0  \\
& Cypriot     & 1800 BCE -- 1600 BCE                 & 60  & 24 \\
& Maltese     & 200 BCE -- 1 BCE                     & 43  & 40 \\
\cline{2-5}
& \multicolumn{2}{r}{\textbf{Subtotal}} & \textbf{888} & \textbf{476} \\

\midrule
\multirow{4}{*}{Bust}
& Greek       & 600 BCE -- 400 BCE                   & 47  & 39 \\
& Roman       & 100 BCE -- 300 CE                    & 44  & 27 \\
& Estonian    & 1900 CE                           & 75  & 2  \\
& German      & 1800 CE -- 2000 CE                   & 52  & 2  \\
\cline{2-5}
& \multicolumn{2}{r}{\textbf{Subtotal}} & \textbf{218} & \textbf{70} \\

\midrule
\multirow{4}{*}{Arrowhead}
& Argentinian & Multiple Periods                 & 94  & 25 \\
& Egyptian    & Neolithic                        & 45  & 12 \\
& Greek       & Ottoman                          & 69  & 7  \\
& Iranian     & 1500 BCE -- 1000 BCE                 & 51  & 1  \\
\cline{2-5}
& \multicolumn{2}{r}{\textbf{Subtotal}} & \textbf{259} & \textbf{45} \\

\midrule
\multirow{6}{*}{Tablet \& Stele}
& Iraqi        & 2100 BCE -- 1000 BCE                 & 54  & 17 \\
& Greek        & Multiple Periods                 & 49  & 27 \\
& Mesopotamian & Bronze Age                      & 174 & 46 \\
& Egyptian     & Multiple Periods                 & 91  & 28 \\
& Roman        & Multiple Periods                 & 13  & 10 \\
\cline{2-5}
& \multicolumn{2}{r}{\textbf{Subtotal}} & \textbf{381} & \textbf{128} \\

\midrule
\multicolumn{3}{r}{\textbf{Total}} & \textbf{1746} & \textbf{719} \\
\bottomrule
\end{tabular}
\caption{Cultural and temporal diversity of the CH damage dataset. The table reports object categories, cultural contexts, associated periods, total object counts, and the number of damaged artifacts annotated with pixel-level segmentation masks.}
\label{tab:ch-dataset-statistics}
\vspace{-20pt}
\end{table*}

We curated all artifacts from Europeana~\footnote{\url{https://europeana.eu/}} or the SHREC2021 dataset~\cite{sipiran2021shrec}. For SHREC2021, we rendered 2D RGB views of the 3D models. We selected objects to include multiple cultural contexts and ensure legal compliance. Only items released under research-compatible licenses (e.g., CC0, CC BY-NC-ND, CC BY-NC-SA) were included. Each record identifies the source repository, cultural attribution, historical period, and material when available. 

Since the images are curated from heterogeneous CH repositories, they were not acquired through a single standardized capture protocol. Acquisition conditions therefore vary in lighting, resolution, viewpoint, object scale, and background characteristics. To reduce background-related variability, we remove the image background and retain the foreground artifact region, so the model focuses on the object itself (visible damage and missing parts). For \emph{DASCH-Net}, we resize all images and corresponding masks to $448 \times 448$ pixels during preprocessing.

\begin{figure}[t!]
    \centering
    \includegraphics[width=0.65\textwidth]{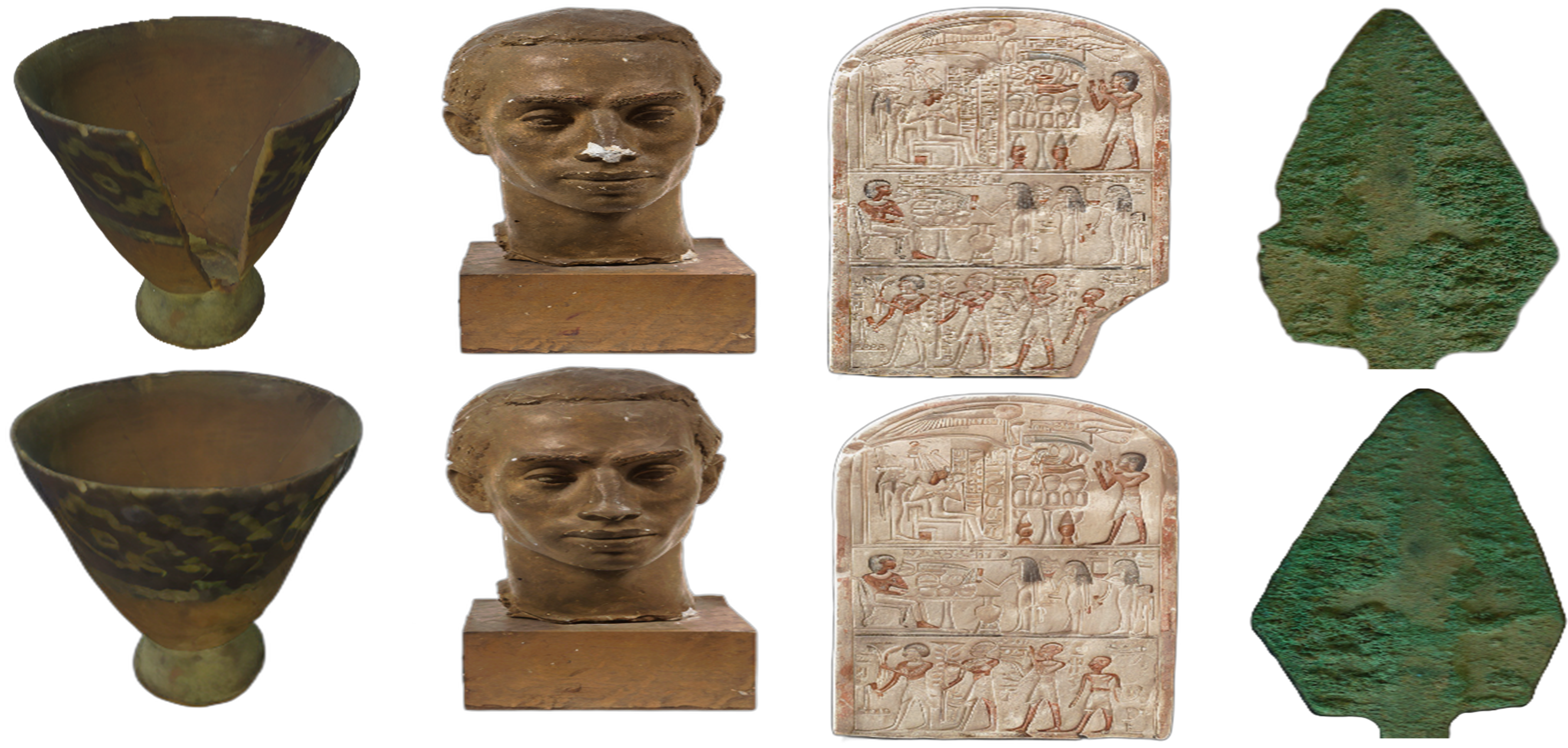}
    \Description{Figure illustrating dataset augmentation via image inpainting for CH artifacts. Each row shows one object category, with the damaged artifact on the left and the corresponding inpainted version on the right. Examples include a ceramic vase with a reconstructed rim, a bust with restored facial regions, an inscribed stone tablet with completed missing parts, and a metallic arrowhead with reconstructed surface continuity. The inpainted images aim to approximate visually plausible intact appearances while preserving the overall object geometry and material characteristics.}
    
    \caption{Examples of inpainting-based augmentation across object categories. Top: damaged objects from the curated dataset. Bottom: inpainted version generated by the authors using Stable Diffusion v1.5.}
    \vspace{-5pt}
    \label{fig:examples-dataset-inpainting}
\end{figure}

\textbf{Data Augmentation.} To mitigate the limited availability of annotated training data, we apply geometric transformations (rotation, scaling, translation, and horizontal and vertical flips), erosion and dilation to introduce variability in damage boundaries and object silhouettes, and synthetic damage augmentation to generate plausible missing-part patterns. To also increase the diversity of intact appearances, we apply image inpainting using Stable Diffusion v1.5~\footnote{\url{https://huggingface.co/stable-diffusion-v1-5/stable-diffusion-v1-5}}. For each object category, we augment $\sim$15\% of samples with an inpainted version conditioned on the original RGB image and the ground-truth damage mask, generating visually plausible intact artifacts (Figure~\ref{fig:examples-dataset-inpainting}).

\textbf{Limitations.} While the dataset provides a diverse and carefully annotated benchmark, limitations remain. The augmentation strategies increase training diversity but do not eliminate the need for larger annotated CH datasets. Its size is modest, reflecting the difficulty in obtaining pixel-level annotation. A single annotator produces the damage masks, ensuring consistency but not capturing inter-annotator variability. The selected object categories also cover only part of CH diversity since Europeana aggregates collections selected for publication by individual contributing institutions, resulting in uneven coverage of cultural memory~\cite{zaagsma2023digital}. Our dataset inherits this source-selection limitation, although we have not quantified collection-level representation. The learned shape priors and damage cues may therefore favor the geometries, materials, and imaging conditions represented in the training data. Thus, reliability on underrepresented collections or different acquisition conditions requires further validation.

\section{Experiments}
\label{sec:experiments}
We conduct (i) quantitative segmentation evaluation, (ii) qualitative analysis through visual overlays, (iii) interpretability assessment via Grad-CAM, and (iv) ablation and exploratory analyses. We aim to assess segmentation accuracy, shape-aware reasoning, interpretability, and robustness under realistic low-resource CH conditions.


\textbf{Quantitative evaluation and baseline comparison.} We compare \emph{DASCH-Net} with representative state-of-the-art segmentation methods, namely Swin-Unet~\cite{cao2022swin}, SegFormer~\cite{xie2021segformer}, DeepLabV3+~\cite{chen2018encoder}, and SegNeXt~\cite{guo2022segnext}. All methods are evaluated on the same artifact categories and test splits. We report three complementary metrics selected to reflect the requirements of CH damage analysis. The Structure Measure ($S_\alpha$)~\cite{fan2017structure} evaluates object-level structural similarity between predicted and ground-truth masks, which is important when damage affects the artifact's silhouette or spatial organization. The Enhanced-alignment Measure ($E_\phi$)~\cite{fan2018enhanced} captures both local pixel alignment and global distribution consistency, making it useful for sparse, fragmented, or irregular damage regions. The mean Dice coefficient (mDice)~\cite{dice1945measures} provides a strict overlap-based measure of pixel-level agreement with human annotations, which is relevant to downstream tasks that require localized masks, such as inpainting and 3D reconstruction. Together, these metrics assess structural coherence, spatial alignment, and segmentation overlap (Section~\ref{subSec:results-quantitative}).

\textbf{Qualitative analysis of damage segmentation.} We analyze representative predictions across object categories. We visualize the input image, ground-truth damage mask, predicted mask, and an overlay on the RGB image (Section~\ref{subSec:results-qualitative}). 

\textbf{Interpretability analysis via Grad-CAM.} To support process-level transparency, we analyze the two main VFM-based components of \emph{DASCH-Net}: the SAM2-based shape estimation model and the DINOv2-based surface damage detector (Section~\ref{subSec:results-interpretability}). For shape estimation, Grad-CAM is computed on the final decoder feature map of the U-Net architecture that produces the silhouette prediction. For surface damage detection, Grad-CAM is applied to the adapter-modified DINOv2 feature maps used for patch-level damage classification.

\textbf{Ablation studies and exploratory analysis.} We evaluate a surface-only variant of \emph{DASCH-Net} in which the shape estimation output is disabled, and damage detection relies only on visible appearance (Section~\ref{subSec:results-ablation}). We also perform cross-category exploratory experiments, in which \emph{DASCH-Net} is trained on one object category and evaluated on unseen categories without fine-tuning.

\textbf{Implementation Details.} All experiments are conducted using PyTorch~\footnote{\url{https://pytorch.org/}} on a single NVIDIA GeForce RTX 4090 Laptop GPU with 16GB of memory and 32GB of system RAM. The VFMs (SAM2~\footnote{\url{https://huggingface.co/docs/transformers/en/model\_doc/sam2}} and DINOv2~\footnote{\url{https://huggingface.co/docs/transformers/en/model\_doc/dinov2}}) are initialized with public pretrained weights. We employ AdamW as optimizer, using an initial learning rate of $1\times10^{-4}$ and a weight decay of $5\times10^{-4}$. Training is performed for 40 epochs with a batch size of 8. A cosine annealing learning rate scheduler is employed throughout training, with a minimum learning rate of $1\times10^{-7}$ to ensure stable convergence. All input images are resized to $448 \times 448$ pixels. We use the PyTorch Grad-CAM package~\footnote{\url{https://github.com/jacobgil/pytorch-grad-cam}} for the interpretability implementation.  

\section{Results \& Discussion}
\label{sec:results-discussion}
\subsection{Quantitative Evaluation of Segmentation Accuracy}
\label{subSec:results-quantitative}
Table~\ref{tab:results-quantitative-results} reports the quantitative comparison between \emph{DASCH-Net} and four state-of-the-art segmentation methods across the four artifact categories. \emph{DASCH-Net} achieves the best performance in 11 out of 12 category-metric combinations. The only exception is $E_\phi$ for the \textbf{Bust} category, where SegNeXt obtains the highest score, while \emph{DASCH-Net} still achieves the best $S_\alpha$ and mDice. Overall, these results indicate that the proposed shape-conditioned formulation improves damage segmentation beyond standard appearance-driven segmentation baselines.

The advantage of \emph{DASCH-Net} is especially visible in categories where damage localization benefits from structural context. Standard segmentation methods rely primarily on RGB appearance and learned texture/semantic cues. This makes them effective when damage has clear visual contrast, but less reliable when missing parts require reasoning about the object's expected extent. By estimating plausible object shape and using inferred missing regions to guide prediction, \emph{DASCH-Net} constrains damage masks to geometrically meaningful regions. This helps explain its consistent improvements over Swin-Unet, SegFormer, DeepLabV3+, and SegNeXt.

Performance also varies substantially across artifact categories. \textbf{Vase \& Bowl} and \textbf{Bust} are the two strongest categories for \emph{DASCH-Net}, although for different reasons. Vase \& Bowl benefits from the largest number of training examples and from relatively regular damage patterns, often occurring around rims, edges, or object bodies. This supports stable learning and accurate boundary localization, leading to the highest $ E_\phi$ score for \emph{DASCH-Net}. Bust achieves the highest $S_\alpha$ and mDice for \emph{DASCH-Net} despite having fewer examples, because damage often appears in semantically consistent facial regions, such as the nose, mouth, or facial contours. These repeated spatial and semantic cues make the damage more learnable than in categories where missing regions are less regular.

For the \textbf{Bust} category, SegNeXt achieves a higher $E_\phi$ than \emph{DASCH-Net}. Since $E_\phi$ measures alignment between the predicted and ground-truth maps by combining local pixel agreement with global distribution consistency, this result suggests that SegNeXt produces predictions that are well aligned with the dominant damaged facial regions. However, its lower $S_\alpha$ and mDice indicate that this alignment does not translate into the best structural consistency or pixel-level overlap. In this case, \emph{DASCH-Net} provides better overall mask quality, while SegNeXt captures the global alignment pattern particularly well.

\begin{table*}[t!]
\centering
\setlength{\tabcolsep}{4pt}
\resizebox{\textwidth}{!}{%
\begin{tabular}{lcccccccccccc}
\toprule
& \multicolumn{3}{c}{\textbf{Vase \& Bowl}} 
& \multicolumn{3}{c}{\textbf{Bust}} 
& \multicolumn{3}{c}{\textbf{Tablet \& Stele}} 
& \multicolumn{3}{c}{\textbf{Arrowhead}} \\
\cmidrule(lr){2-4} \cmidrule(lr){5-7} \cmidrule(lr){8-10} \cmidrule(lr){11-13}
\textbf{Method} 
& $\mathbf{S_\alpha}$ & $\mathbf{E_\phi}$ & mDice
& $\mathbf{S_\alpha}$ & $\mathbf{E_\phi}$ & mDice
& $\mathbf{S_\alpha}$ & $\mathbf{E_\phi}$ & mDice
& $\mathbf{S_\alpha}$ & $\mathbf{E_\phi}$ & mDice \\
\midrule
Swin-Unet~\cite{cao2022swin}                 & 70.4 & 71.7 & 52.2 & 59.8 & 58.2 & 28.0 & 56.4 & 68.3 & 37.9 & 52.0 & 30.4 & 4.4 \\
SegFormer~\cite{xie2021segformer}              & 79.2 & 85.7 & 69.5 & 80.9 & 86.7 & 64.2 & 59.8 & 68.3 & 46.1 & 51.9 & 25.0 & 0.0 \\
DeepLabV3+~\cite{chen2018encoder}        & 80.0 & 89.5 & 72.1 & 74.4 & 76.1 & 51.4 & 58.7 & 63.9 & 43.2 & 51.9 & 25.0 & 0.0 \\
SegNeXt~\cite{guo2022segnext}     & 78.8 & 87.7 & 70.3 & 79.5 & \textbf{92.4} & 65.9 & 57.1 & 62.3 & 39.2 & 51.9 & 25.0 & 0.0 \\
Ours          & \textbf{81.0} & \textbf{92.1} & \textbf{73.6} & \textbf{88.3} & 89.1 & \textbf{74.9} & \textbf{60.9} & \textbf{69.3} & \textbf{48.8} & \textbf{59.7} & \textbf{66.5} & \textbf{27.7} \\
\bottomrule
\end{tabular}%
}
\caption{Quantitative segmentation results per artifact category, comparing DASCH-Net with different segmentation approaches. All metrics are reported as percentages (\%), with higher values indicating better performance. For Arrowhead, SegFormer, DeepLabV3+, and SegNeXt produced empty damage masks, yielding the same scores (51.9/25.0/0.0).}
\label{tab:results-quantitative-results}
\vspace{-18pt}
\end{table*}

\textbf{Tablet \& Stele} remains more challenging across methods. Damage in this category often consists of large, irregular, and fragmented missing regions with ambiguous boundaries. As a result, structurally plausible predictions may still be penalized by mDice when their extent does not precisely match the human annotation. Nevertheless, \emph{DASCH-Net} achieves the best scores across all three metrics, suggesting that shape information provides useful context even when exact pixel-level agreement is difficult. \textbf{Arrowhead} is the most difficult category because damage is sparse, fine-grained, and highly imbalanced relative to the visible object area. Most baseline methods collapse toward background-dominated predictions, as reflected by near-zero mDice values. The substantially higher $E_\phi$ and mDice obtained by \emph{DASCH-Net} indicate that shape-conditioned reasoning is particularly useful when visual evidence alone is weak or underrepresented.

These results highlight the importance of complementary metrics. mDice captures strict pixel-level overlap, while $S_\alpha$ and $E_\phi$ better reflect structural coherence and alignment. In CH damage analysis, this distinction is important because a prediction may be structurally meaningful for downstream tasks such as inpainting or 3D reconstruction even when its exact pixel-level extent differs from the annotation.

\subsection{Qualitative Analysis of Damage Segmentation}
\label{subSec:results-qualitative}
\begin{figure}[t!]
    \centering

    \newcommand{\qualimage}{%
    \marginbox{0pt 0pt 0pt .006\height}{%
            \includegraphics[width=\linewidth]%
                {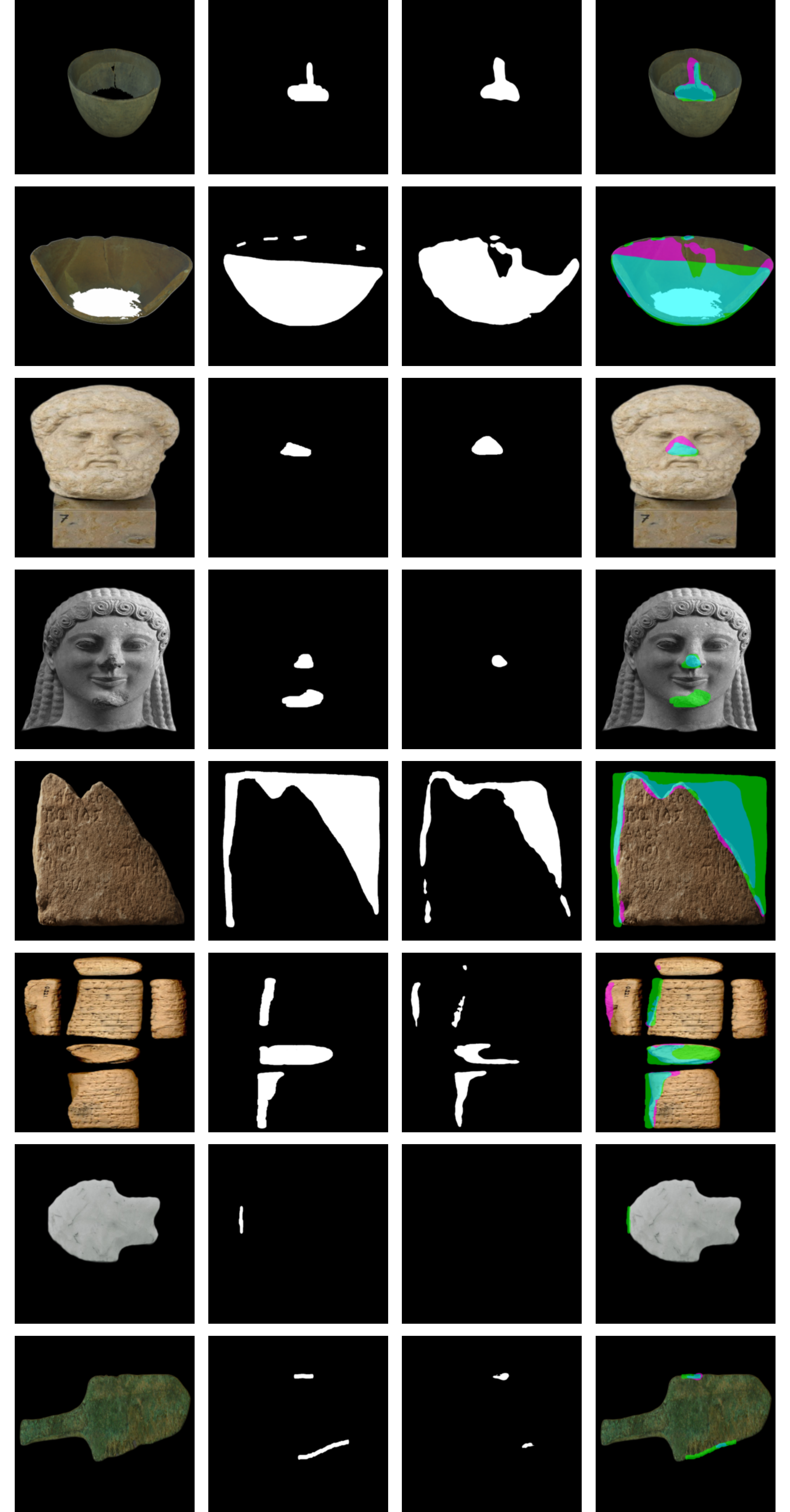}%
        }%
    }
    
    \newcommand{\qualheadings}{%
        \noindent
        \makebox[.25\linewidth][c]{\strut Image}%
        \makebox[.25\linewidth][c]{GT}%
        \makebox[.25\linewidth][c]{Ours}%
        \makebox[.25\linewidth][c]{GT vs Ours}%
    }
    
    \newcommand{\qualblock}[2]{%
        \begin{minipage}[t]{.40\linewidth}
            \centering
            {\large\qualheadings\par}
            \noindent
            \clipbox{0pt {#1\height} 0pt {#2\height}}{\qualimage}%
        \end{minipage}%
    }
    
    \qualblock{.5}{0}\hspace{10mm}\qualblock{0}{.5}

    \Description{Qualitative comparison of damage segmentation results across the four categories of CH artifacts considered in this work. Each row corresponds to a different artifact. Columns show the original RGB image, the ground-truth damage mask, the prediction produced by DASCH-Net, and an overlay of ground-truth and predicted damage regions on the original image.}

    \caption{Qualitative results across the four categories of CH artifacts. From left to right, each row shows the original RGB image, the ground-truth damage mask, the DASCH-Net predicted damage mask, and an overlay of the GT and prediction on the original image. In the overlay, GT regions are shown in {\textcolor{green}{green}}, predicted regions in {\textcolor{magenta}{magenta}}, and overlap is in {\textcolor{cyan}{cyan}}.}
    \label{fig:results-qualitative-comparison}
\end{figure}

The qualitative results in Figure~\ref{fig:results-qualitative-comparison} illustrate typical success cases and category-specific challenges. The \textbf{Vase \& Bowl} predictions generally align well with ground truth, capturing both the location and the structure of damaged regions. Minor spatial shifts or over-extensions mainly affect mDice, while structural metrics remain high. We can see an important failure case in the second example. The damage corresponds to a large missing area on the object's front surface, and the prediction fails to capture its full extent. This behavior can be attributed to the object's hollow geometry: visual features extracted from the RGB image may originate from intact interior or backside surfaces, producing cues that resemble undamaged exterior regions. In such cases, single-view RGB evidence and silhouette-level shape cues are insufficient to distinguish missing exterior surfaces from visible interior surfaces. Without explicit depth, surface-layer, or multi-view/3D constraints, the model misses surface-level reasoning. We argue that resolving such ambiguities may require a solution that explicitly separates global shape completeness from local surface integrity. While existing depth-aware and amodal shape representations do not directly address this at the pixel level for CH artifacts, they suggest promising directions for adaptation and future investigation in this context~\cite{nadeem2025maskadapt}.

For \textbf{Bust}, the model reliably localizes damage in frequent regions such as the nose, yielding structurally coherent predictions despite occasional boundary imprecision. Less frequent damage locations (e.g., chin) are sometimes missed, reflecting intra-category imbalance in damage distribution. The analysis for \textbf{Tablet \& Stele} shows that the model captures the overall location and structure of damage but often underestimates its full extent, particularly for large or fragmented missing parts. Partial recovery and false positives in visually ambiguous areas explain the reduced overlap-based scores. The \textbf{Arrowhead} results reflect the difficulty of detecting small and sparse damage on thin structures. The model may either miss damage entirely or recover only part of the affected region, leading to strong performance penalization. 

Finally, across categories, performance depends strongly on damage scale, geometry, and spatial regularity. Structurally coherent predictions do not always yield high overlap scores, reinforcing the need to interpret quantitative metrics in the context of object-specific damage characteristics.

\subsection{Interpretability Analysis via Grad-CAM}
\label{subSec:results-interpretability}
\begin{figure}[t!]
    \centering

        \newcommand{\shapecamblock}[1]{%
        \begin{minipage}[t]{0.45\linewidth}
            \centering
            {\large
                \centering
                \noindent
                \makebox[.20\linewidth][c]{\strut Image}%
                \makebox[.20\linewidth][c]{Original}%
                \makebox[.20\linewidth][c]{GT}%
                \makebox[.20\linewidth][c]{Ours}%
                \hspace*{-.04\linewidth}%
                \makebox[.26\linewidth][c]{\normalsize Grad-CAM}%
                \par
            }
            \noindent
            \clipbox{#1}{%
                \includegraphics[width=\linewidth]%
                    {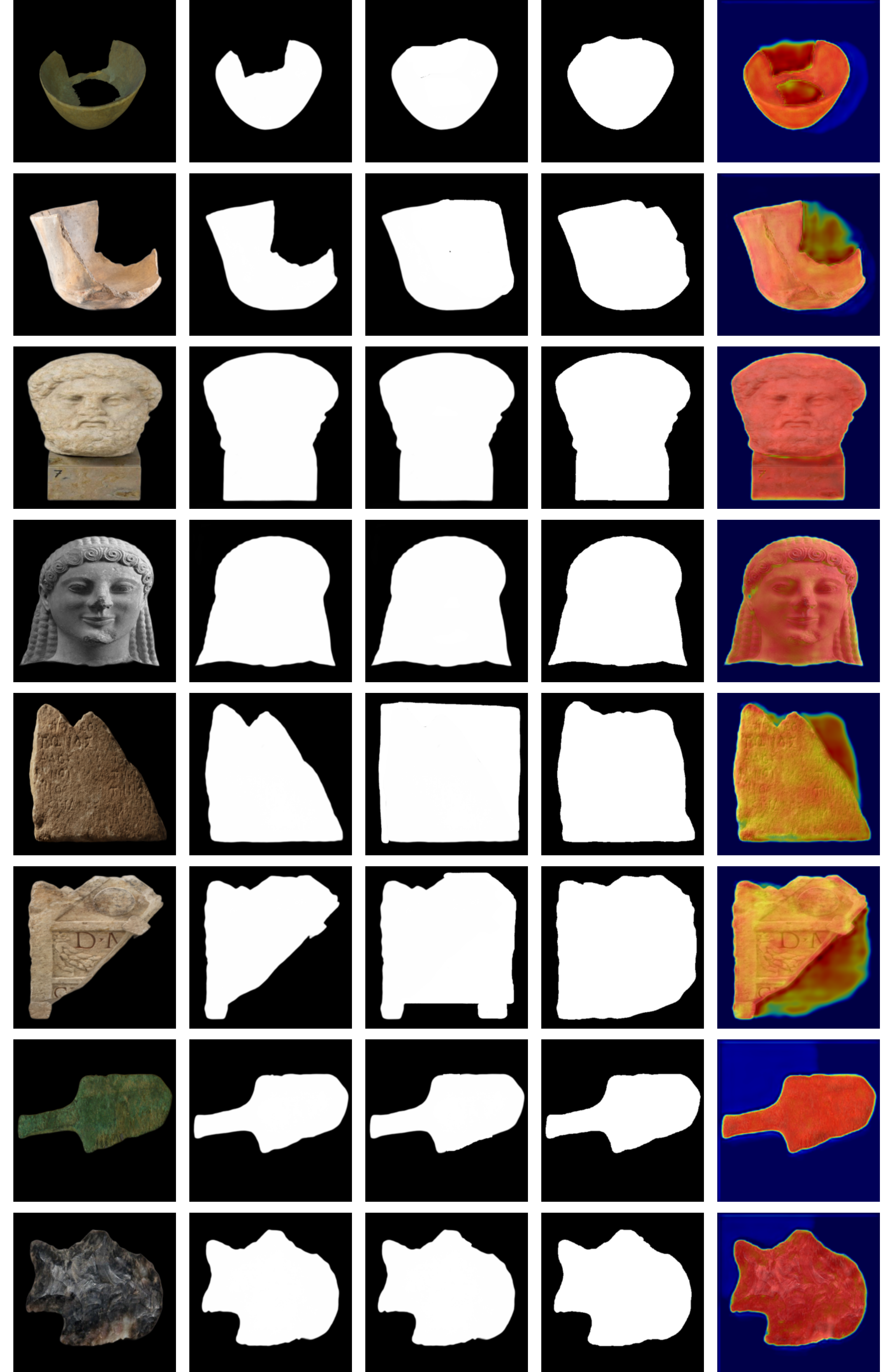}%
            }%
        \end{minipage}%
    }

    \shapecamblock{0pt {.5\height} 0pt 0pt}%
    \hspace{10mm}%
    \shapecamblock{0pt 0pt 0pt {.5\height}}
    
    \Description{Grad-CAM–based interpretability results for the shape prior component of DASCH-Net. Each row corresponds to a different CH artifact. Columns show the original RGB image, the original visible object shape, the ground-truth complete shape, the predicted complete shape, and the Grad-CAM attribution map, which indicates the spatial regions that most influence the shape completion.}

    \caption{Grad-CAM–based interpretability analysis of the shape prior component. From left to right, each row shows the original RGB image, the original visible shape, the ground-truth (GT) complete shape, the predicted complete shape produced by DASCH-Net, and the corresponding Grad-CAM attribution map.}
    \label{fig:results-shape-prior-interpretability}
\end{figure}

\textbf{Missing-Part and Shape Estimation.} Grad-CAM maps (Figure~\ref{fig:results-shape-prior-interpretability}) for the shape estimation component show strong activation over the global object silhouette and along fracture parts, particularly in regions requiring missing-part inference. Background areas remain largely inactive, indicating that shape completion is driven by geometrically meaningful visual cues rather than contextual bias. Across artifact categories, the attribution patterns consistently align with object structure and damaged boundaries, supporting the conclusion that shape-completion predictions rely on structural evidence relevant to missing-part reasoning.

\begin{figure}[t!]
    \centering
    \newcommand{\surfacecamblock}[1]{%
        \begin{minipage}[t]{0.40\linewidth}
            \centering
            {\large
                \noindent
                \makebox[.25\linewidth][c]{\strut Image}%
                \makebox[.25\linewidth][c]{GT}%
                \makebox[.25\linewidth][c]{Ours}%
                \makebox[.25\linewidth][c]{Grad-CAM}%
                \par
            }
            \noindent
            \clipbox{#1}{%
                \includegraphics[width=\linewidth]{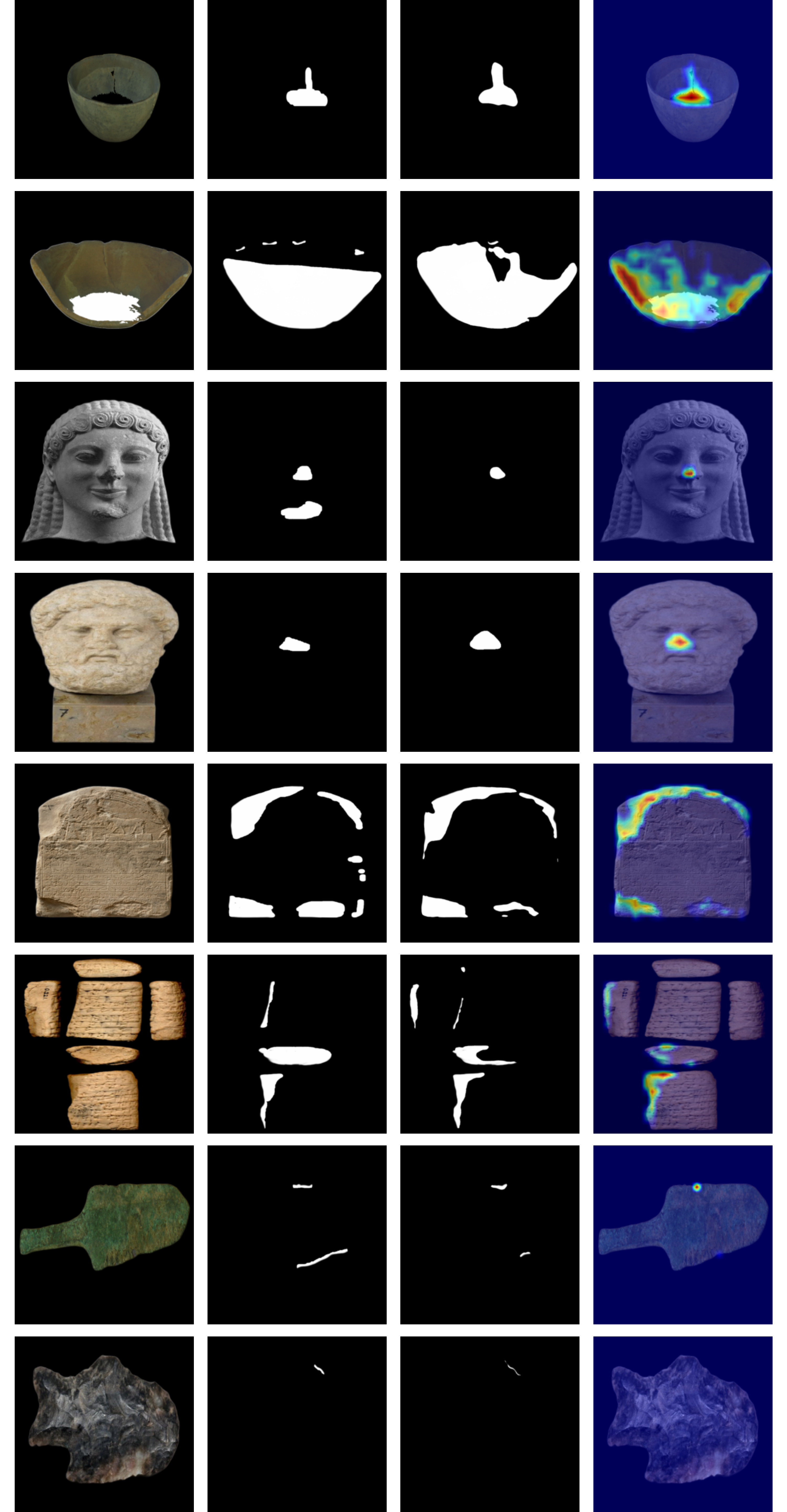}%
            }%
        \end{minipage}%
    }

    \surfacecamblock{0pt {.5\height} 0pt 0pt}%
    \hspace{10mm}%
    \surfacecamblock{0pt 0pt 0pt {.5\height}}
    
    \Description{Grad-CAM–based interpretability results for the surface damage detection component of DASCH-Net. Each row corresponds to a different CH artifact. Columns show the original RGB image, the ground-truth surface damage mask, the predicted damage mask, and the Grad-CAM attribution map highlighting image regions that most influence the damage prediction.}
    
    \caption{Grad-CAM–based interpretability analysis of the surface damage detection component. From left to right, each row shows the original RGB image, the ground-truth (GT) surface damage mask, the predicted damage mask produced by DASCH-Net, and the corresponding Grad-CAM attribution map. 
    }
    \vspace{-10pt}
    \label{fig:results-surface-damage-interpretability}
\end{figure}

\textbf{Surface Damage Evidence.} Grad-CAM activations (Figure~\ref{fig:results-surface-damage-interpretability}) concentrate on visually degraded regions, such as erosion or texture loss, and align closely with predicted damage masks. For undamaged artifacts, activation is minimal, supporting the idea that damage predictions by this component are driven by surface-level visual evidence. In cases of diffuse degradation (Tablet \& Stele), broader activation patterns reflect sensitivity to extended surface irregularities.

\subsection{Ablation Studies and Exploratory Analysis}
\label{subSec:results-ablation}
\textbf{Shape-Prior Relevance.} Table~\ref{tab:ablation-full-vs-shape} compares the full shape-aware \emph{DASCH-Net} with a variant in which the shape estimation output is not used for predicting the damage mask. Results highlight how the contribution of shape reasoning depends on damage characteristics. For \textbf{Vase \& Bowl} and \textbf{Bust}, incorporating shape priors consistently improves performance, particularly for localizing small or spatially constrained damaged regions. This indicates that geometric context enhances robustness beyond appearance cues alone, although the surface-only variant still captures useful visual evidence in these categories. For \textbf{Tablet \& Stele}, removing shape priors also leads to a clear performance drop, especially in mDice, showing that structural context remains beneficial even in this more difficult category. This category remains challenging because damage often appears as large, irregular, and fragmented missing regions, where small deviations in predicted extent are strongly penalized by pixel-level overlap metrics. Finally, performance in the \textbf{Arrowhead} category collapses under surface-only reasoning, underscoring the critical role of shape context when damage is sparse, fine-grained, and severely imbalanced. 

\begin{table}[t!]
\centering
\begin{tabular}{lcccccc}
\toprule
& \multicolumn{2}{c}{$S_\alpha$} & \multicolumn{2}{c}{$E_\phi$} & \multicolumn{2}{c}{mDice} \\
Category 
& Surface (\%) & $\Delta$ (pp) 
& Surface (\%) & $\Delta$ (pp) 
& Surface (\%) & $\Delta$ (pp) \\
\midrule
Vase \& Bowl     
& 76.5 & $-4.5$ 
& 86.0 & $-6.1$ 
& 65.7 & $-7.9$ \\

Bust   
& 81.0 & $-7.3$ 
& 80.6 & $-8.5$ 
& 64.2 & $-10.7$ \\

Tablet \& Stele  
& 53.0 & $-7.9$ 
& 59.7 & $-9.6$ 
& 33.0 & $-15.8$ \\

Arrowhead        
& 52.0 & $-7.7$ 
& 25.5 & $-41.0$ 
& 1.0  & $-26.7$ \\
\bottomrule
\end{tabular}
\caption{Surface-only damage detection results and absolute performance differences ($\Delta$) expressed in percentage points (pp) relative to the full DASCH-Net results reported in Table~\ref{tab:results-quantitative-results}. Negative values indicate performance degradation when shape priors are not used.}
\label{tab:ablation-full-vs-shape}
\vspace{-15pt}
\end{table}

\textbf{Cross-Category Generalization.} Exploratory experiments show that both shape estimation and damage localization degrade substantially under category shift. Models trained on a single category fail to generalize reliably to unseen object types with different geometries and materials. While this limits direct cross-category deployment, it reflects the intrinsic variability of CH artifacts. As no consistent quantitative trends emerged across categories, we present representative failure cases qualitatively in the supplementary material.

\textbf{Use Scenario.} A customs officer may encounter a \textit{potentially} trafficked CH object, such as a damaged vase with incomplete form that complicates identification. The officer would photograph the object and send the RGB image to a CH expert for evaluation using DASCH-Net. The expert could inspect the original object, estimated complete shape, inferred missing regions, and final damage mask to assess the object's condition and guide subsequent digital restoration and 3D generation. In practice, these outputs provide additional visual information for documenting the object and evaluating its cultural relevance. Retaining the original photograph and explicitly marking inferred or repaired regions keeps observed material distinguishable from reconstruction hypotheses during expert assessment and comparison with collection records.

\textbf{Additional Applicability.} DASCH-Net's surface-only variant (Table~\ref{tab:ablation-full-vs-shape}) supports damage localization without silhouette completion, suggesting potential for analyzing surface erosion on buildings or cracks and paint loss in paintings. Such damage may share visual cues with degradation on tablets and steles, although differences in materials and acquisition conditions would require domain training and validation. Inferring missing structural parts would also require retraining the shape-prior component. In paintings, this prior could help identify missing canvas or panel edges.

\section{Conclusion \& Future Work}
\label{sec:conclusion}
This paper introduced \emph{DASCH-Net} (Section~\ref{sec:dasch-net}), a hierarchical shape-informed framework for damage analysis of Cultural Heritage (CH) artifacts from a single RGB image. We formulated damage analysis as a joint problem of intact shape estimation and surface-level damage detection. By explicitly coupling these tasks through a shape prior, \emph{DASCH-Net} enables reasoning about damaged objects whose original geometry is only partially observable. \emph{DASCH-Net} combines a learned VAE prior with SAM2 to reason about missing regions beyond visible evidence (Section~\ref{subSec:shape-prior-component}), while surface damage detection leverages DINOv2 representations adapted via lightweight modules (Section~\ref{subSec:surface-damage-component}). This modular design encourages geometric consistency and exposes intermediate predictions for inspection.

We also introduced a curated dataset of movable CH artifacts with pixel-level annotations of surface degradation and missing parts (Section~\ref{sec:dataset}). The dataset covers diverse object categories, cultural contexts, historical periods, materials, and damage patterns, providing a first benchmark for studying damage analysis in movable CH artifacts where both visible surface degradation and structural loss must be considered. Experiments combining quantitative metrics, qualitative analysis, diagnostic visualization, and ablation studies demonstrate that \emph{DASCH-Net} provides a strong reference framework for damage analysis and segmentation in low-resource CH settings (Section~\ref{sec:results-discussion}). In comparison with state-of-the-art segmentation methods (Swin-Unet, SegFormer, DeepLabV3+, and SegNeXt), \emph{DASCH-Net} achieves the best performance in most category-metric combinations, with particularly clear gains when damage localization requires structural context. These results support the value of conditioning damage prediction on estimated object shape, especially for sparse, irregular, or missing-part damage. Finally, through process-level transparency and Grad-CAM-based diagnostic visualization, results indicate that \emph{DASCH-Net} relies on task-appropriate evidence: global structural cues guide shape estimation, while localized appearance cues support surface damage detection. The intermediate outputs (estimated shape, inferred missing-region cues, and final damage masks) make the prediction process more traceable for conservation, documentation, and digital reconstruction workflows.

\textbf{Limitations}. The framework operates on single-view RGB images and therefore cannot resolve depth ordering or internal surface ambiguity. Cross-category generalization also remains limited, reflecting the dependence of shape priors and damage semantics on object geometry and material properties. In addition, while the proposed dataset covers a range of artifact types and damage patterns, broader coverage is required to better reflect the diversity of CH collections.

\textbf{Future Work} will explore several directions. First, we plan to integrate the predicted damage masks into downstream pipelines for digital restoration (e.g., image inpainting, 3D reconstruction, scale-consistent shape generation). This could lead to a long-term goal of producing 3D-printable components for conservation and restoration. Second, we will extend \emph{DASCH-Net} to multi-view and 3D settings, enabling more reliable reasoning about object depth, surface continuity, and internal structure. Third, we plan to investigate domain-adaptive and few-shot learning strategies to improve cross-category generalization. Finally, we envision incorporating expert-in-the-loop feedback and multi-annotator studies to better capture diverse conservation perspectives and to further align automated damage analysis with real-world CH practice.

\section*{Acknowledgments}
This research was performed under the EU ENIGMA project (Grant Agreement No. 101094237), and it has received funding from the Swiss State Secretariat for Education, Research and Innovation (SERI), contract number 22.00467, ref-1131-52104. We also thank our project partners for their valuable discussions and insights that contributed to the development of this work.

\bibliographystyle{ACM-Reference-Format}
\bibliography{references}

\end{document}